\documentclass[11pt]{article}
\usepackage[T1]{fontenc}
\usepackage[utf8]{inputenc}
\usepackage{lmodern}
\usepackage[margin=1in,headheight=30pt,headsep=12pt]{geometry}
\usepackage{amsmath,amssymb,amsthm,graphicx}
\usepackage{booktabs}
\usepackage{enumitem}

\usepackage[sort,compress]{cite}
\usepackage{xcolor}
\usepackage{tikz}
\usepackage{pgfplots}
\pgfplotsset{compat=1.18}
\usepackage{algorithm}
\usepackage{algpseudocode}
\usepackage{fontawesome}
\usepackage{subcaption}
\usepackage{bibunits}
\usepackage[verbose,hypertexnames=false]{hyperref}
\hypersetup{
  hidelinks,
  pdftitle={Data, Numbers, and Geometry: Three Tutorials on Computation, Learning, and Evaluation},
  pdfauthor={Jessica N. Howard, Yidi Qi, Tom\'as S. R. Silva}
}
\usepackage{fancyhdr}
\newcommand{\TopTutorialRunningTitle}{}
\newcommand{\TopTutorialRunningAuthor}{}
\fancypagestyle{tutorialpages}{%
  \fancyhf{}%
  \fancyhead[C]{%
    \normalfont\small\itshape
    \makebox[\headwidth][s]{\TopTutorialRunningTitle\hfill\TopTutorialRunningAuthor}%
  }%
  \fancyfoot[C]{\thepage}%
}
\numberwithin{equation}{section}
\newtheorem{theorem}{Theorem}[section]
\theoremstyle{definition}
\newtheorem{definition}{Definition}[section]
\newenvironment{arabiclist}
  {\begin{enumerate}[label=\arabic*.,leftmargin=*]}
  {\end{enumerate}}
\newenvironment{romanlist}[1][(iv)]
  {\begin{enumerate}[label=(\roman*),widest=iv,leftmargin=*]}
  {\end{enumerate}}
\newcommand{\tbl}[2]{\centering\caption{#1}{\small #2}\par}
\let\colrule\midrule
\let\botrule\bottomrule
\providecommand{\alttext}[1]{}
\newcommand{\catchline}[5]{}
\newenvironment{history}{}{}
\newcommand{\received}[1]{}
\newcommand{\revised}[1]{}
\newcommand{\accepted}[1]{}
\newcommand{\published}[1]{}
\newcommand{\address}[1]{}
\newcommand{\keywords}[1]{\par\smallskip{\small\noindent\textbf{Keywords:} #1\par}}
\newcommand{\ccode}[1]{\par\smallskip{\small\noindent #1\par}}

\newcommand{\rd}{\mathrm{d}}
\newcommand{\Dflux}{\mathrm{D}}
\newcommand{\RR}{\mathbb{R}}
\newcommand{\vol}{\operatorname{vol}}
\DeclareMathOperator{\curl}{curl}
\DeclareMathOperator{\grad}{grad}
\DeclareMathOperator{\dive}{div}
\newcommand{\QQ}{\mathbb{Q}}
\newcommand{\bh}{\mathbf{h}}
\newcommand{\bX}{\mathbf{X}}
\newcommand{\bH}{\mathbf{H}}
\newcommand{\ap}{\tilde{a}}
\DeclareMathOperator{\ReLU}{ReLU}
\DeclareMathOperator{\MLP}{MLP}

\newcommand{\background}[0]{\text{A}}
\newcommand{\signal}[0]{\text{B}}

\makeatletter
\AtBeginDocument{%
  \let\TopLabel\label
  \let\TopRef\ref
  \let\TopPageRef\pageref
  \let\TopCiteNumbers\@cite@n
}
\newif\ifTopTutorialTitlePrinted
\newenvironment{tutorial}[5]{%
  \clearpage
  \renewcommand{\TopTutorialRunningTitle}{#3}%
  \renewcommand{\TopTutorialRunningAuthor}{#2}%
  \pagestyle{tutorialpages}%
  \TopTutorialTitlePrintedfalse
  \let\TopTutorialAddress\@empty
  \def\TopTutorialHeading{%
    \thispagestyle{plain}%
    \par
    \begingroup
      \centering
      {\normalfont\Large\bfseries Tutorial #1: #3\par}%
      \vspace{1em}%
      {\normalfont #2\par}%
      \ifx\TopTutorialAddress\@empty\else
        \vspace{0.4em}%
        {\small\itshape\TopTutorialAddress\par}%
      \fi
    \endgroup
    \medskip
    \TopTutorialTitlePrintedtrue
  }%
  \def\title##1{}%
  \def\author##1{}%
  \renewcommand{\address}[1]{\def\TopTutorialAddress{##1}}%
  \def\maketitle{%
    \ifTopTutorialTitlePrinted\else\TopTutorialHeading\fi
  }%
  \setcounter{section}{0}%
  \setcounter{subsection}{0}%
  \setcounter{subsubsection}{0}%
  \setcounter{equation}{0}%
  \setcounter{figure}{0}%
  \setcounter{table}{0}%
  \setcounter{algorithm}{0}%
  \setcounter{footnote}{0}%
  \graphicspath{{#4/}}%
  \renewcommand{\label}[1]{\TopLabel{tutorial#1:##1}}%
  \let\ltx@label\label % amsmath uses this entry point for equation labels.
  \renewcommand{\ref}[1]{\TopRef{tutorial#1:##1}}%
  \renewcommand{\pageref}[1]{\TopPageRef{tutorial#1:##1}}%
  \pdfbookmark[0]{Tutorial #1: #3}{tutorial-#1}%
  \begin{bibunit}[ws-ijdsms]%
  \renewcommand{\@cite@n}[1]{%
    \@startbibunitorrelax
    \begingroup
      \let\@auxout\@bibunitaux
      \TopCiteNumbers{##1}%
    \endgroup
  }%
  \renewcommand{\bibliographystyle}[1]{}%
  \renewcommand{\bibliography}[1]{\putbib[#5]}%
}{%
  \ifTopTutorialTitlePrinted\else\TopTutorialHeading\fi
  \clearpage
  \end{bibunit}%
}
\makeatother

\begin{document}

\title{Data, Numbers, and Geometry:\\
Three Tutorials on Numerical Methods, Machine Learning, and Evaluation}

\author{Jessica N. Howard, Yidi Qi, and Tom\'as S. R. Silva}
\date{}

\maketitle

\pdfbookmark[0]{Introduction}{general-introduction}
\section*{\centering Introduction}

These notes bring together three technique tutorials from \emph{DANGER: Data, Numbers, and Geometry}, held at the Banff International Research Station in April 2026. They offer a practical introduction to computing with mathematical data and using machine learning in mathematical research. Our emphasis is on the choices behind a calculation: how the data are represented, which structure a method uses, and how its results are assessed.

Tom\'as S. R. Silva begins with exterior calculus when a differential form is available only through pointwise evaluations. Expressing the exterior derivative through boundary fluxes leads to a numerical method. Examples in Euclidean space and on the sphere show how geometric identities and topological features can be explored from data, with attention to approximation errors and finite precision.

Yidi Qi examines how the structure of mathematical data guides the choice of a neural network. The examples involve elliptic curves, quivers, and a boundary value problem. They show how architectural choices affect learning and which properties can be built into a model. The final example uses interval arithmetic to bound the residual of a trained network over the full interval of the boundary value problem.

Jessica N. Howard discusses how to evaluate and present machine learning results. The tutorial moves from choosing performance metrics and estimating uncertainty to designing clear, accessible figures. Classification serves as a detailed example, showing how decision thresholds and receiver operating characteristic curves help assess a model’s performance.

Throughout the collection, numerical agreement, predictive accuracy, and rigorous bounds are treated as different kinds of evidence. Each tutorial can be read independently, with its own notation and references. We encourage readers to work through the accompanying notebooks and exercises, vary the examples, and try the methods on questions of their own.

%=========================================================================
% TUTORIAL 1 -- Tom\'as S. R. Silva
% Source: first_tutorial_Tomas/main.tex
% The document body from this source is embedded unchanged below.
%=========================================================================
\begin{tutorial}{1}{Tom\'as S. R. Silva}
  {A data-driven view of exterior calculus}
  {.}{references}

% >>> BEGIN PASTE: TUTORIAL 1 DOCUMENT BODY (UNCHANGED) >>>

\markboth{T. S. R. Silva}{A Data-Driven View of Exterior Calculus}

%%%%%%%%%%%%%%%%%%%%% Publisher's Area please ignore %%%%%%%%%%%%%%%%%%%%%%%%
\catchline{}{}{}{}{}
%%%%%%%%%%%%%%%%%%%%%%%%%%%%%%%%%%%%%%%%%%%%%%%%%%%%%%%%%%%%%%%%%%%%%%%%%%%%

\title{A data-driven view of exterior calculus}

\author{Tom\'as S. R. Silva}

\address{Instituto de Matem\'atica, Estat\'istica e Computa\c{c}\~ao
Cient\'ifica (IMECC),\\
Universidade Estadual de Campinas (UNICAMP), Rua S\'ergio Buarque de
Holanda, 651,\\
13083-859 Campinas, S\~ao Paulo, Brazil\\
tomas@ime.unicamp.br}

\maketitle

\begin{history}
\received{(Day Month Year)}
\revised{(Day Month Year)}
\accepted{(Day Month Year)}
\published{(Day Month Year)}
\end{history}

\begin{abstract}
Classical computational tools for differential forms assume symbolic
expressions, mesh discretizations, or smoothness close to
$C^\infty$. In data-driven settings one often has only \emph{oracle
access}: the ability to evaluate a form at arbitrary points and
tangent vectors, with no closed-form expression and no mesh. Building
on a flux-based reformulation of the exterior derivative, recently
developed by the author with D. Fadel and H. N. S\'a Earp, in which
the exterior derivative is recovered as a limit of normalized
boundary integrals over shrinking blocks, we describe a mesh-free,
symbol-free numerical scheme for exterior differentiation, valid
under regularity assumptions weaker than $C^1$. We demonstrate the
method in $\RR$, $\RR^3$ and on the sphere $S^2$, verifying Stokes'
theorem, the identity $\rd^2=0$, and Gauss--Bonnet purely from
pointwise samples, and detecting de Rham cohomology
(closed-but-not-exact forms) from data. Companion Jupyter notebooks
accompany the paper. This article is based on a tutorial delivered at
\emph{DANGER: Data, Numbers, and Geometry} (Banff, April 2026).
\end{abstract}

\keywords{Exterior derivative; Stokes' theorem; numerical exterior
derivative; black-box methods; differential forms; de Rham
cohomology.}

\ccode{Mathematics Subject Classification 2020: 26B05, 58A10, 65D25}

%============================================================================
\section{Motivation: The Black-Box Perspective}
\label{sec:motivation}

Differential forms are the common language for flux through surfaces,
circulation along curves, densities, and conservation laws; they
organize electromagnetism, fluid vorticity, curvature in general
relativity, and, increasingly, data analysis on manifolds. The
unifying operation is the exterior derivative $\rd$, which subsumes
gradient, curl and divergence into a single operator satisfying
$\rd^2=0$ and encoding topology through the obstructions to solving
$\rd\eta=\omega$.

Textbook methods for \emph{computing} with forms, however, assume too
much. They require a symbolic formula for the form (often
unavailable), a mesh or triangulation of the domain (expensive in
high dimension), or smoothness close to $C^\infty$ (too restrictive).
Discrete Exterior Calculus computes $\rd$ via cochains on a
simplicial mesh~\cite{Hirani}, and Finite Element Exterior Calculus
exploits the structure of a discrete de Rham
complex~\cite{ArnoldFEEC}; both are powerful, and both presuppose a
discretized domain. In practice -- for instance when the form is
produced by a trained neural network, a simulation, or a measurement
process -- we often have only \emph{pointwise samples}: ``We can
evaluate $\omega_p(v_1,\ldots,v_k)$ for any point $p$ and tangent
vectors $v_i$, but we have no closed-form expression and no mesh.''

This paper takes that constraint as the starting point. Given only
oracle access to $\omega$:
\begin{arabiclist}
\item Can we compute $\rd\omega$ \textbf{numerically}?
\item Can we \textbf{verify} Stokes' theorem from data?
\item Can we \textbf{detect topology} (e.g., closed but not exact)
  from data?
\end{arabiclist}
The answer to all three is yes, and the key insight is a geometric
reading of the derivative, made rigorous in Ref.~\citen{FSS}.
Sections \ref{sec:flux} and \ref{sec:algorithm} summarize the theory
and the resulting algorithm; Secs.~\ref{sec:flat} and
\ref{sec:sphere} report numerical experiments in flat space and on
the sphere, answering questions 1--3 in turn.

%============================================================================
\section{The Exterior Derivative as Infinitesimal Flux}
\label{sec:flux}

Start in dimension one. If $f:\RR\to\RR$ is differentiable at $a$,
then
\begin{equation}
f'(a) \;=\; \lim_{r\to 0}\frac{f(a+r)-f(a-r)}{2r}
        \;=\; \lim_{r\to 0}\frac{1}{\vol\bigl(B_r(a)\bigr)}
              \int_{\partial B_r(a)} f ,
\label{eq:1d}
\end{equation}
where the interval $B_r(a)=[a-r,a+r]$ is regarded as an oriented
one-chain, so that $\int_{\partial B_r(a)}f = f(a+r)-f(a-r)$. Read
geometrically: \emph{the derivative is the infinitesimal flux of $f$
through the boundary of a shrinking ball}, normalized by volume. This
viewpoint, first proposed by
Acker\cite{Acker1996,Acker2004,Acker2012} and developed
systematically in Ref.~\citen{FSS}, generalizes verbatim to
$(k-1)$-forms in $\RR^n$.

\begin{definition}[Flux-based exterior derivative, Ref.~\citen{FSS},
Def.~11]
\label{def:D}
Let $\omega$ be a $(k-1)$-form on an open set $U\subset\RR^n$,
flux-integrable around $x\in U$. We say $\omega$ is \emph{derivable}
at $x$, with exterior derivative
$\Dflux\omega_x\in\Lambda^k(\RR^n)^\ast$, if for every open set
$A\subset\RR^k$, every point $p\in A$, every choice of vectors
$v_1,\dots,v_k\in\RR^n$, and every $C^1$ map $\varphi:A\to U$ with
$\varphi(p)=x$ and $\varphi'(p)e_i=v_i$ for $i=1,\dots,k$,
\begin{equation}
\Dflux\omega_x(v_1,\dots,v_k)
\;=\;
\lim_{B\to p}\;\frac{1}{\vol(B)}\int_{\partial B}\varphi^\ast\omega ,
\label{eq:D}
\end{equation}
where, for every fixed $K>1$, the limit is taken over $k$-blocks
$B\subset A$ containing $p$ and shrinking to $p$, with
$L(B)/l(B)<K$.
\end{definition}

The aspect-ratio condition controls
$\operatorname{diam}(B)\cdot\operatorname{area}(\partial B)/\vol(B)$
and is what makes the definition well posed; see Sec.~3 of
Ref.~\citen{FSS}. Crucially, derivability is \emph{strictly weaker}
than classical differentiability -- $\omega$ need not even be
continuous at $x$ for $\Dflux\omega_x$ to exist. The main structural
results are the following.

\begin{theorem}[Ref.~\citen{FSS}, Secs.~2--4]
\label{thm:main}
Let $U\subset\RR^n$ be open.
\begin{romanlist}[(iv)]
\item \textbf{(Mean value equality)} If $\omega$ is a flux-continuous
$(k-1)$-form on a $k$-block $B$, derivable on $\mathring{B}$, then
there exists $\xi\in\mathring{B}$ with
$\Dflux\omega_\xi(e_1,\dots,e_k)=\frac{1}{\vol(B)}\int_{\partial
B}\omega$.
\item \textbf{(Stokes without $C^1$)} If $\omega\in\Omega^{k}(U)$ is
continuous on $c(B)$ and differentiable on $c(\mathring{B})$ for a
$C^1$ singular $(k+1)$-block $c:B\to U$, then $\int_{\partial
c}\omega=\int_{c}\rd\omega$ whenever the right-hand side exists.
\item \textbf{($\Dflux^2=0$)} Under natural integrability hypotheses,
$\Dflux(\Dflux\omega)=0$.
\item \textbf{(Compatibility)} If $\omega$ is differentiable at
$x_0$, then $\omega$ is derivable at $x_0$ and
$\Dflux\omega_{x_0}=\rd\omega_{x_0}$.
\end{romanlist}
\end{theorem}

Item (i) is a genuine higher-dimensional Mean Value Theorem \emph{in
equality form}, obtained by iterating a Trisection Lemma on nested
sub-blocks; item (ii) then follows by the same subdivision argument
that derives the Fundamental Theorem of Calculus from the classical
MVT, with the usual $C^1$ hypothesis on the integrand relaxed to
exactly the FTC-type assumptions. For the present purposes the
punchline is item (iv): Eq.~(\ref{eq:D}) computes the classical
$\rd\omega$ whenever the latter exists. \emph{The flux formula is
therefore a licensed numerical definition of the exterior
derivative} -- and it consumes only pointwise evaluations
of~$\omega$.

%============================================================================
\section{The Black-Box Algorithm and its Error Budget}
\label{sec:algorithm}

Fix a point $x\in\RR^n$, a degree $k$, and a step size
$\varepsilon>0$. Discretizing Eq.~(\ref{eq:D}) on the axis-aligned
cube of side $2\varepsilon$ spanned by $e_{q_1},\dots,e_{q_k}$, and
approximating each face integral by its midpoint value, yields the
centered stencil\cite{FSS}
\begin{equation}
\Dflux\omega_x(e_{q_1},\dots,e_{q_k})
\;\approx\;
\frac{1}{2\varepsilon}
\sum_{i=1}^{k}\sum_{j=0}^{1}
(-1)^{i+j}\,
\omega_{\,x+(-1)^{j+1}\varepsilon e_{q_i}}
\bigl(e_{q_1},\dots,\widehat{e_{q_i}},\dots,e_{q_k}\bigr),
\label{eq:stencil}
\end{equation}
for each increasing multi-index $Q=\{q_1<\dots<q_k\}$. The scheme is
summarized in Algorithm~\ref{alg:blackbox}; a reference
implementation is available at
\begin{center}
\url{https://github.com/TomasSilva/NumericalExteriorDerivative}. 
\end{center}

\begin{algorithm}[!ht]
\caption{Numerical exterior derivative in $\RR^n$ (black-box)}
\label{alg:blackbox}
\begin{algorithmic}[1]
\Require oracle \textsc{sampler} for a $(k-1)$-form $\omega$; point
$x\in\RR^n$; step $\varepsilon>0$
\Ensure numerical components of $\Dflux\omega_x$
\State initialize the component array of
  $\Dflux\omega_x\in\Lambda^k(\RR^n)^\ast$ to zero
\For{all multi-indices $q_1<\dots<q_k$}
  \State set $\Dflux\omega_x(e_{q_1},\ldots,e_{q_k})$ equal to the
  right-hand side of Eq.~(\ref{eq:stencil}), using $2k$ oracle queries
\EndFor
\State extend the increasing-index components to all indices by
  antisymmetry
\State \Return $\Dflux\omega_x$
\end{algorithmic}
\end{algorithm}

\subsection{Properties}
No mesh is required -- only pointwise
evaluations. No symbolic expression is needed -- the form is a pure
black box. The underlying flux definition remains meaningful under
regularity assumptions weaker than $C^1$ (Theorem~\ref{thm:main}).
For oracle forms with locally $C^3$ coefficient functions, the
centered stencil has truncation error $O(\varepsilon^2)$.

\subsection{The Fundamental Tension}
\label{sec:error-budget}
In floating-point arithmetic the
truncation error $O(\varepsilon^2)$ competes against catastrophic
cancellation $O(\varepsilon_{\rm mach}/\varepsilon)$ in the
difference quotient. The total error is V-shaped in $\varepsilon$
(Fig.~\ref{fig:vcurve}), with sweet spot
$\varepsilon^\ast\sim\varepsilon_{\rm mach}^{1/3}\approx 10^{-5}$ in
double precision. When the oracle is \emph{noisy} with amplitude
$\sigma$ -- the relevant regime when $\omega$ is produced by a
trained network or by measurements -- the optimum shifts to
$\varepsilon^\ast\sim\sigma^{1/3}$: noisier oracles demand
\emph{larger} probe regions.

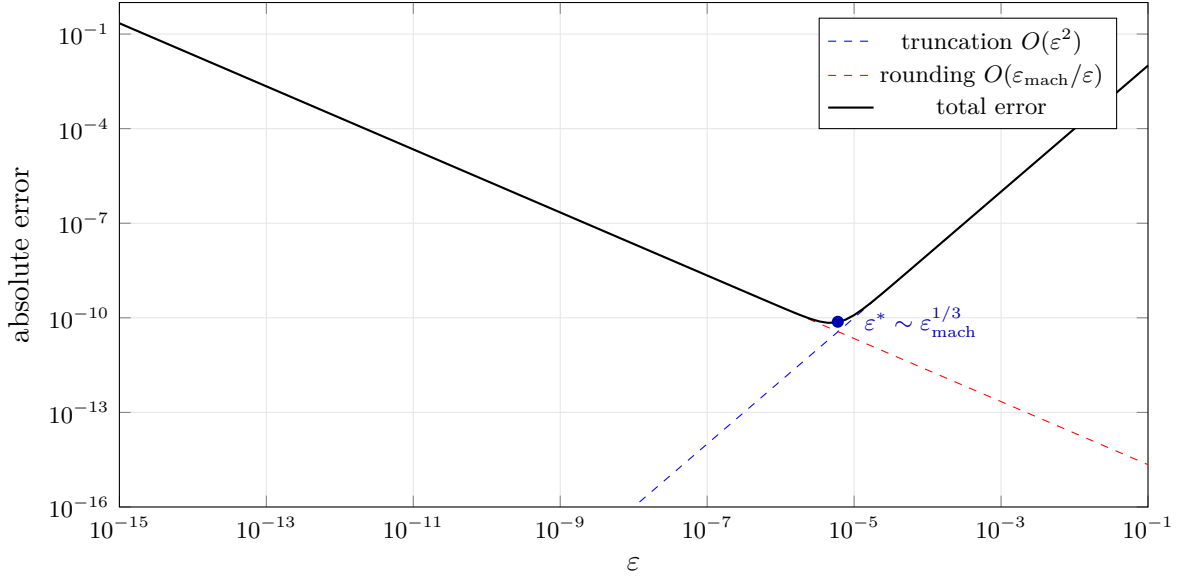
\begin{figure}[!ht]
\centerline{%
\begin{tikzpicture}
\begin{loglogaxis}[
  width=0.92\textwidth, height=0.5\textwidth,
  xlabel={$\varepsilon$}, ylabel={absolute error},
  xmin=1e-15, xmax=1e-1, ymin=1e-16, ymax=1e0,
  xtick={1e-15,1e-13,1e-11,1e-9,1e-7,1e-5,1e-3,1e-1},
  tick label style={font=\footnotesize},
  legend pos=north east, legend style={font=\footnotesize},
  grid=major, grid style={black!10}]
\addplot[blue, dashed, domain=1e-15:1e-1, samples=60] {x^2};
\addlegendentry{truncation $O(\varepsilon^2)$}
\addplot[red, dashed, domain=1e-15:1e-1, samples=60] {2.2e-16/x};
\addlegendentry{rounding $O(\varepsilon_{\rm mach}/\varepsilon)$}
\addplot[black, thick, domain=1e-15:1e-1, samples=120]
  {x^2 + 2.2e-16/x};
\addlegendentry{total error}
\node[circle,fill=blue!70!black,inner sep=1.6pt]
  at (axis cs:6e-6,7.5e-11) {};
\node[anchor=west, blue!70!black, font=\footnotesize]
  at (axis cs:1e-5,7.5e-11)
  {$\varepsilon^\ast\sim\varepsilon_{\rm mach}^{1/3}$};
\end{loglogaxis}
\end{tikzpicture}}
% \vspace*{8pt}
\caption{Truncation vs. cancellation for the centered stencil of
Eq.~(\protect\ref{eq:stencil}). The optimum
$\varepsilon^\ast\approx10^{-5}$ balances the two regimes; under
oracle noise of amplitude $\sigma$ it shifts to
$\varepsilon^\ast\sim\sigma^{1/3}$.}
\label{fig:vcurve}
\alttext{Log-log plot of absolute error against step size, showing a
V-shaped total error curve formed by a decreasing rounding branch and
an increasing truncation branch, with the optimum near ten to the
minus five.}
\end{figure}

%============================================================================
\section{Experiments in Flat Space}
\label{sec:flat}

In $\RR^3$ the single formula (\ref{eq:stencil}) specializes to the
three classical operators of vector calculus: the probe region grows
with the degree, and only the dimension of the boundary changes
(Table~\ref{tab:unified}). All that is needed is black-box evaluation
of the form: a zero-form is an oracle
$\texttt{f(point)}\to\texttt{float}$, a one-form returns the
component vector $[\omega(e_1),\omega(e_2),\omega(e_3)]$, a two-form
returns the skew-symmetric matrix $[\eta(e_i,e_j)]$.

\begin{table}[!ht]
\tbl{The unified picture in $\RR^3$: one algorithm, three classical
operators.\label{tab:unified}}
{\begin{tabular}{@{}lccc@{}}
\toprule
 & 0-form $f$ & 1-form $\omega$ & 2-form $\eta$ \\
\colrule
output of $\rd$      & 1-form $\rd f$      & 2-form $\rd\omega$   & 3-form $\rd\eta$ \\
classical name       & gradient            & curl                 & divergence \\
probe region         & segment $[a,b]$     & parallelogram        & parallelepiped \\
boundary $\partial$  & 2 oriented points   & 4 oriented edges     & 6 oriented faces \\[2pt]
boundary integral    & $f(b)-f(a)$
  & $\displaystyle\sum_{\rm edges}\int\omega$
  & $\displaystyle\sum_{\rm faces}\iint\eta$ \\[10pt]
$\rd^2=0$ says\ldots & $\curl(\grad f)=0$  & $\dive(\curl\omega)=0$ & \emph{(top degree)} \\
\botrule
\end{tabular}}
\end{table}

\subsection{Exact and Non-closed Forms}
For
$\omega = x\,\rd x + y\,\rd y + z\,\rd z =
\rd\bigl(\tfrac12(x^2{+}y^2{+}z^2)\bigr)$ the algorithm returns
$\rd\omega=0$ to machine precision. For the non-closed form
$\omega = z\,\rd x + x\,\rd y + y\,\rd z$ it recovers
$\curl(\omega)=(1,1,1)$ with error $\sim10^{-11}$.

\subsection{$\rd^2=0$ from Data}
Starting from
$f(x,y,z)=e^{-(x^2+y^2+z^2)}$, we compute $\rd f$ numerically (a
one-form), then $\rd(\rd f)$ numerically (a two-form), chaining two
applications of Algorithm~\ref{alg:blackbox}. The result is
$|\rd^2 f|\sim10^{-17}$: the identity $\curl(\grad f)=0$ is confirmed
from data alone.

\subsection{Topology from Data: The Vortex Form}
Consider
\begin{equation}
\omega \;=\; \frac{-y\,\rd x + x\,\rd y}{x^2+y^2}
\qquad\text{on } \RR^3\setminus\{z\text{-axis}\}.
\label{eq:vortex}
\end{equation}
The algorithm sees: $\rd\omega=0$ at every test point away from the
axis (the form is \emph{closed}), and $|\rd\omega|\to\infty$ when the
stencil straddles the singular axis. What the algorithm \emph{cannot}
see pointwise: there is no function $f$ with $\rd f=\omega$; the form
is \emph{not exact}, and
$\oint_\gamma\omega=2\pi\,\operatorname{wind}
(\gamma,\text{$z$-axis})$; in particular, it equals $2\pi$ for a
positively oriented loop of winding number one. The method recovers
this value by numerical line integration of oracle samples. The
lesson: $\rd\omega=0$ is
\emph{local} (checkable pointwise by Algorithm~\ref{alg:blackbox}),
while non-exactness is \emph{global} -- it requires integrating
around loops. The gap between the two is de Rham cohomology, and it
is detectable from data.

\subsection{Stokes' Theorem from Data}
Take
$\omega=z\,\rd x+x\,\rd y+y\,\rd z$, $\Sigma$ the unit disk in the
plane $z=1$, $\partial\Sigma$ the unit circle. Computing
$\int_\Sigma\rd\omega$ by applying Algorithm~\ref{alg:blackbox} at
quadrature points and $\oint_{\partial\Sigma}\omega$ by direct oracle
evaluation along the boundary gives
\begin{equation}
\int_\Sigma \rd\omega \;=\; 3.14159265\ldots\,,
\qquad
\oint_{\partial\Sigma}\omega \;=\; 3.14159265\ldots\,,
\qquad
|{\rm diff}|\sim 10^{-8},
\label{eq:stokesR3}
\end{equation}
with both sides computed from black-box evaluations only -- no
formula for $\curl(\omega)$ was used.

%============================================================================
\section{Experiments in Curved Manifolds: The Sphere}
\label{sec:sphere}

On $S^2\subset\RR^3$ three things change: the tangent space
$T_pS^2=\{v: v\cdot p=0\}$ varies with $p$; stencil points
$p\pm\varepsilon t$ leave the sphere; and a two-form is top degree
($\dim S^2=2$), hence a scalar density. The fix is minimal: replace
the canonical basis $e_i$ by an orthonormal tangent frame $(t_1,t_2)$
at $p$, and project stencil points radially back to the sphere,
$\pi(q)=q/|q|$. Set
$p_{\pm i}=\pi(p\pm\varepsilon t_i)$ and
$s_\varepsilon=(1+\varepsilon^2)^{-1/2}$. For zero- and one-forms,
respectively,
\begin{equation}
\begin{aligned}
(\rd f)_p(t_i)
&\;\approx\;\frac{f(p_{+i})-f(p_{-i})}{2\varepsilon},\\
(\rd\omega)_p(t_1,t_2)
&\;\approx\;\frac{s_\varepsilon}{2\varepsilon}
 \bigl[\omega_{p_{+1}}(t_2)-\omega_{p_{-1}}(t_2)\\
&\hphantom{\;\approx\;\frac{s_\varepsilon}{2\varepsilon}\bigl[}
 {}-\omega_{p_{+2}}(t_1)+\omega_{p_{-2}}(t_1)\bigr].
\end{aligned}
\label{eq:sphere-stencil}
\end{equation}
The factor $s_\varepsilon$ is the tangent-vector scaling induced by
the differential of the radial projection. The structural insight
making this work is naturality. For the local parametrization
$\Phi(u_1,u_2)=\pi(p+u_1t_1+u_2t_2)$, classical naturality gives
$\rd(\Phi^\ast\omega)=\Phi^\ast(\rd\omega)$, so the flat-space
stencil applied to $\Phi^\ast\omega$ computes the intrinsic exterior
derivative at $p$. The corresponding naturality statement for
$\Dflux$ on open Euclidean sets is Ref.~\citen{FSS}, Lemma~13. For
sufficiently smooth oracle data, this chart-based stencil retains
$O(\varepsilon^2)$ convergence and the same
$\varepsilon_{\rm mach}^{1/3}$ scaling of the optimal step size.

\subsection{Gradient of the Height Function}
For $f(p)=z$ the
intrinsic gradient $\nabla_{S^2}f$ should point toward the north
pole, have magnitude $\sin\theta$, and vanish at the poles. The
algorithm recovers $|\nabla_{S^2}f|=\sin\theta$ to the displayed six
decimal places at the equator ($1.000000$), at $45^\circ$N
($0.707107$), near the pole ($0.010000$) and at the pole itself
($0.000000$), locating the critical points exactly.

\subsection{\texorpdfstring{$\rd\omega$}{dw} Matches the Analytical Result}
For the
rotation form $\omega=x\,\rd y-y\,\rd x$ pulled back to $S^2$, the
analytical answer is
$\rd\omega|_{S^2}=2z\cdot\omega_{\rm area}$. Over 200 random points
the maximal error is $\sim10^{-10}$; the computed density is positive
in the northern hemisphere, negative in the southern, and vanishes at
the equator. The algorithm sees the geometry of the sphere --
curvature, latitude -- through black-box evaluations alone.

\subsection{Stokes on Geodesic Caps}
For the cap
$\Sigma=\{z\ge z_0\}$ with boundary circle $\partial\Sigma$ at height
$z_0$, both sides of
$\int_\Sigma\rd\omega=\oint_{\partial\Sigma}\omega$ equal
$2\pi(1-z_0^2)$ exactly. The numerical verification is reported in
Table~\ref{tab:caps}: neither side used a formula for $\rd\omega$.

\begin{table}[!ht]
\tbl{Stokes' theorem on geodesic caps $\Sigma=\{z\ge z_0\}\subset
S^2$, verified from black-box evaluations only. Exact value:
$2\pi(1-z_0^2)$.\label{tab:caps}}
{\begin{tabular}{@{}cccc@{}}
\toprule
$z_0$ & surface integral & line integral & $|{\rm diff}|$ \\
\colrule
0.0 & 6.2831853 & 6.2831853 & $\sim10^{-8}$ \\
0.5 & 4.7123890 & 4.7123890 & $\sim10^{-8}$ \\
0.9 & 1.1938052 & 1.1938052 & $\sim10^{-8}$ \\
\botrule
\end{tabular}}
\end{table}

\subsection{Gauss--Bonnet}
Integrating the area form
$\omega_{\rm area}(t_1,t_2)=p\cdot(t_1\times t_2)$ over the whole
sphere by oracle sampling recovers
$\int_{S^2}\omega_{\rm area}=4\pi$. Since the unit sphere has
$K\equiv1$, this area integral also equals the Gauss--Bonnet
total-curvature integral
$\int_{S^2}K\,\rd A=2\pi\chi(S^2)=4\pi$, with $\chi(S^2)=2$.
It is the total curvature $\int K\,\rd A$, rather than the area
itself, that is topologically invariant.

\subsection{$\rd^2=0$ on a Curved Manifold}
Chaining two numerical
derivatives on the sphere -- $f(p)=z$, then $\rd f$, then
$\rd(\rd f)$ -- yields $|\rd^2f|\sim10^{-6}$ at every test point.
The error is larger than in $\RR^3$ because two numerical derivatives
are chained, each using projected stencil points. In exact arithmetic
it converges to zero as $\varepsilon\to0$; in finite precision,
convergence persists only down to the optimal step size, after which
cancellation dominates. The identity $\rd^2=0$ is universal, and
verifiable from black-box evaluations alone.

\subsection{Summary}
Pointwise evaluation of differential forms is
sufficient to recover their exterior derivatives, verify Stokes'
theorem, and detect topological features -- on flat spaces and
curved manifolds alike.

%============================================================================
\section{Outlook: Learning Geometry from Data}
\label{sec:outlook}

The natural habitat of the black-box scheme is the growing interface
between machine learning and differential geometry, where forms are
increasingly \emph{learned} rather than written down. In
Ref.~\citen{HHSS}, neural networks learn a Ricci-flat metric on a
Calabi--Yau threefold and approximate the ${\rm G}_2$-structure
three-form on the associated seven-dimensional link in $S^9$; the
numerical exterior derivative described here serves as the
\emph{validation layer}, certifying (co)closure conditions of the
learned forms without symbolic access to them. In that regime the
noisy-oracle analysis of Sec.~\ref{sec:error-budget}
($\varepsilon^\ast\sim\sigma^{1/3}$) is the operative one.

Natural next steps include systematic exterior calculus on learned
implicit manifolds, estimation of de Rham cohomology from sampled
forms via the local-to-global gap exhibited by the vortex example,
and higher-degree stencils in the ambient dimensions relevant to
special holonomy. The three tutorial notebooks
(\texttt{block1\_1d\_story.ipynb}, \texttt{block2\_R3\_forms.ipynb},
\texttt{block3\_S2\_forms.ipynb}), covering every experiment reported
above with guided exercises, are available at
\url{https://github.com/TomasSilva/Banff-Tutorial}, alongside the
SageMath\cite{sagemath} reference implementation.

\section*{Acknowledgments}
The author thanks Daniel Fadel and Henrique N. S\'a Earp, co-authors
of the survey \cite{FSS} underlying this tutorial, and the DANGER organizers --
Edward Hirst, Elli Heyes, Miranda Cheng and Sergei Gukov -- for the invitation and
for assembling such a stimulating meeting. The ideas revisited here
originate with Felipe
Acker \cite{Acker1996,Acker2004,Acker2012}. The author was supported
by the S\~ao Paulo Research Foundation (FAPESP), grant 2022/09891-4.

\section*{ORCID}
\noindent Tom\'as S. R. Silva -- \url{https://orcid.org/0000-0002-5825-9133}

\bibliographystyle{ws-ijdsms}
\bibliography{references}

% <<< END PASTE: TUTORIAL 1 DOCUMENT BODY <<<

\end{tutorial}

%=========================================================================
% TUTORIAL 2 -- Yidi Qi
% Source: second_tutorial_Yidi/main.tex
% The document body from this source is embedded unchanged below.
%=========================================================================
\begin{tutorial}{2}{Yidi Qi}
  {Architectures for mathematical data}
  {.}{references}

% >>> BEGIN PASTE: TUTORIAL 2 DOCUMENT BODY (UNCHANGED) >>>

\markboth{Y. Qi}{Architectures for Mathematical Data}

%%%%%%%%%%%%%%%%%%%%% Publisher's Area please ignore %%%%%%%%%%%%%%%%%%%%%%%%
\catchline{}{}{}{}{}
%%%%%%%%%%%%%%%%%%%%%%%%%%%%%%%%%%%%%%%%%%%%%%%%%%%%%%%%%%%%%%%%%%%%%%%%%%%%

\title{Architectures for mathematical data}

\author{Yidi Qi}

\address{Department of Physics, Northeastern University,\\
360 Huntington Avenue, Boston, Massachusetts 02115, USA\\
% TODO: confirm affiliation lines and add/remove the IAIFI line as appropriate
The NSF AI Institute for Artificial Intelligence and Fundamental
Interactions (IAIFI)\\
qi.yid@northeastern.edu}

\maketitle

\begin{history}
\received{(Day Month Year)}
\revised{(Day Month Year)}
\accepted{(Day Month Year)}
\published{(Day Month Year)}
\end{history}

\begin{abstract}
Mathematical objects rarely arrive as unstructured feature vectors. They
arrive as sequences indexed by primes, as directed graphs closed under a
combinatorial move, as functions constrained by a differential equation.
Choosing a neural architecture is therefore not a matter of choosing the
largest model, but of choosing an \emph{inductive bias} that matches the
structure the data already has. We develop this principle through three
self-contained case studies, each implemented from scratch in JAX and each
runnable on a laptop CPU. (i)~\emph{Sequential data}: a 1D CNN and a
single-block transformer predict the rank of an elliptic curve
$E/\QQ$ from its normalized Frobenius traces $\ap_p$; a saliency analysis
tracked across training shows the network localizing the discriminative
signal in the small primes, where the murmuration phenomenon lives.
(ii)~\emph{Graph data}: a directed graph isomorphism network classifies
quivers by Dynkin mutation type, is permutation equivariant to machine zero
by construction, and---trained only on $7$ and $8$ vertices---transfers to
quivers on $9$ vertices. (iii)~\emph{Continuous data}: a physics-informed neural network
solves a two-point boundary value problem, after which interval arithmetic
bounds the residual rigorously over the \emph{entire} domain rather than at the
collocation points. 
Companion Jupyter notebooks can be viewed at https://github.com/yidiq7/Banff-Tutorial-Architecture. 
Based on a tutorial delivered at \emph{DANGER: Data, Numbers, and Geometry} (Banff, April 2026).
\end{abstract}

\keywords{Inductive bias; convolutional neural networks; transformers; graph
neural networks; permutation equivariance; physics-informed neural networks;
interval arithmetic; elliptic curve rank; quiver mutation.}

%============================================================================
\section{Introduction: Inductive biases of the architectures}
\label{sec:introduction}

Machine learning enters mathematics along two broadly distinct routes. In the
first, the model is a \emph{reasoning engine}: language models and agents that
prove theorems, generate conjectures, or search proof spaces. In the second,
mathematical objects are treated as \emph{structured data}---invariants,
sequences, graphs, functions---and the task is to learn a map from that data to
a quantity of interest, sometimes as a route to human
insight.\cite{Davies2021,He2023book} This tutorial is about the second route, and about one
question inside it: \emph{given the data, which
architecture?}

The question has content because architectures are not interchangeable
function approximators. Each carries an \textbf{inductive bias}: structural
assumptions baked into the hypothesis class before training. A convolution
assumes the same local pattern is meaningful wherever it occurs; a
message-passing layer assumes the answer does not depend on how the vertices
are numbered; a physics-informed loss assumes the target satisfies a
differential equation. When the assumption matches the mathematics it buys
sample efficiency, generalization and---as in Sec.~\ref{sec:graph}---\emph{exact}
structural guarantees no amount of training data can supply. 
Table~\ref{tab:landscape} is the map used to organize the tutorial.

\begin{table}[!ht]
\tbl{The landscape: from the structure of the data to the architecture that
encodes it.\label{tab:landscape}}
{\begin{tabular}{@{}p{2.1cm}p{2.5cm}p{3.0cm}p{4.0cm}@{}}
\toprule
Data type & Key structure & Architecture(s) & Inductive bias \\
\colrule
Tabular & unstructured features & MLP, boosted trees & smoothness;
axis-aligned splits \\
Sequential & order, locality & 1D CNN, transformer, SSM & shift
equivariance; attention; recurrence \\
Grid / image & 2D spatial structure & 2D CNN, ViT, U-Net & 2D translation
equivariance, locality \\
Set / multiset & no ordering & DeepSets, set transformer &
permutation-invariant aggregation \\
Graph & nodes, edges, topology & GNN (message passing) & permutation
equivariance, locality \\
Geometric & symmetry group ($E(3)$, \ldots) & equivariant networks & group
equivariance by construction \\
Continuous / PDE & smoothness, differential equations & PINN, FNO, DeepONet &
PDE constraints; operator learning \\
\botrule
\end{tabular}}
\end{table}

We develop three of the rows further below against genuine mathematical problems and datasets:
\emph{sequential} (Sec.~\ref{sec:seq}), predicting the rank of an
elliptic curve over $\QQ$ from its Frobenius traces; \emph{graph}
(Sec.~\ref{sec:graph}), classifying quivers by Dynkin mutation type; and
\emph{continuous} (Sec.~\ref{sec:pinn}), solving a boundary value problem with
a PINN and then \emph{verifying} the result rigorously.

%============================================================================
\section{Sequential Data: Elliptic Curve Rank from Frobenius Traces}
\label{sec:seq}

\subsection{The data and the task}

For an elliptic curve $E/\QQ$ of conductor $N$ and a prime $p\nmid N$, the
Frobenius trace is $a_p(E) = p + 1 - \#E(\mathbb{F}_p)$, and the Hasse bound
$|a_p|\le 2\sqrt{p}$ makes
\begin{equation}
\ap_p(E) \;=\; \frac{a_p(E)}{2\sqrt{p}} \;\in\; [-1,1]
\label{eq:normalize}
\end{equation}
a natural normalization.\cite{Silverman} Collectively the $a_p$ determine
$L(E,s)=\sum_n a_n n^{-s}$, whose order of vanishing at $s=1$ is the analytic
rank; Birch--Swinnerton-Dyer predicts this equals the rank of $E(\QQ)$. Our
dataset consists of $17{,}340$ curves of conductor $N\le 3000$ ($9{,}323$ of
rank $0$, $8{,}017$ of rank $1$), each represented by the vector
$(\ap_2,\ap_3,\dots,\ap_{9973})$ of $1{,}229$ normalized traces at all primes
$p<10^4$. The task is binary classification of the analytic rank, $0$ versus
$1$, on an $80/20$ split
($13{,}872$ train, $3{,}468$ test), with per-position standardization fitted
on the training set only.

What makes this \emph{sequential} rather than tabular is the
\textbf{murmuration} phenomenon: averaging $a_p$ over all curves of a fixed
rank in a fixed conductor range and plotting against $p$ produces an
oscillation whose phase and sign depend on the
rank.\cite{Murmurations,Zubrilina} In the conductor range used here it shows
up as a large excursion of the two class averages, opposite in sign and
concentrated at small $p$ (Fig.~\ref{fig:murm}(a),(b)), which is invisible in
any single curve (Fig.~\ref{fig:murm}(c)). The discriminative structure is
therefore a pattern \emph{along} the prime axis---exactly what a
locality-respecting architecture is built to detect.

\begin{figure}[!ht]
\centerline{\includegraphics[width=\textwidth]{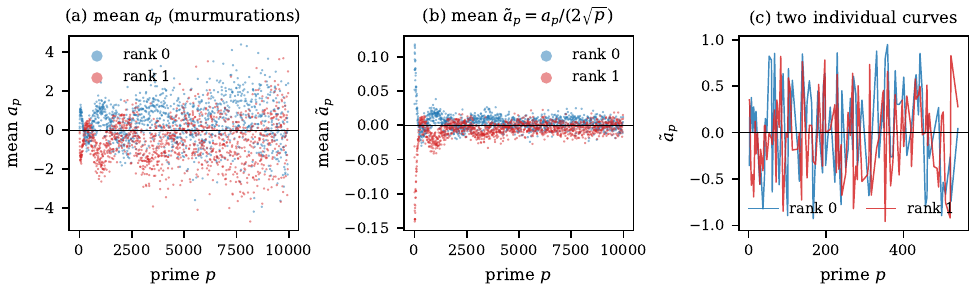}}
\caption{(a) Class-averaged Frobenius traces $a_p$ over the training split.
(b) The same after Hasse normalization, $\ap_p = a_p/(2\sqrt p)$. In both, the
two class averages are near mirror images and their separation is largest at
small $p$. (c) Two individual curves over the first $100$ primes: the
per-curve signal is indistinguishable from noise, so the pattern is a property
of the \emph{sequence}, recoverable only in aggregate.}
\label{fig:murm}
\alttext{Three panels. The first two show class-averaged Frobenius traces
against prime, unnormalized and normalized, with rank zero above and rank one
below the axis and the largest excursions at small primes. The third shows two
individual curves fluctuating apparently randomly.}
\end{figure}

\subsection{Two architectures, one parameter budget}

A 1D convolution slides a kernel of width $k$ along the sequence,
\begin{equation}
(W * x)_t \;=\; \sum_{j=0}^{k-1} W_j\, x_{t+j},
\label{eq:conv}
\end{equation}
so a pattern detected at one position produces the same response
elsewhere.\cite{LeCun1998} This is \emph{translation equivariance} along the
prime index; with pooling, a few kernels can look for local oscillations at
every scale. Our CNN is
$\mathrm{Conv}(1{\to}8,k{=}5)\to\ReLU\to\mathrm{MaxPool}(4)
\to\mathrm{Conv}(8{\to}16,k{=}5)\to\ReLU\to\mathrm{MaxPool}(4)$, flattened to
$76\times 16$ and classified by a two-layer head.

The control is an MLP of shape $1{,}229\to 32\to 16\to 2$, which sees the input as
a flat vector in $\RR^{1229}$: every prime position gets dedicated weights and
the ordering of the primes is, to the model, arbitrary. The two models are
sized to essentially the same parameter count ($39{,}714$ versus $39{,}922$),
so the comparison is about inductive bias rather than capacity, and both are
trained identically: Adam, learning rate $10^{-3}$, batch size $512$, ten
epochs.

\begin{table}[!ht]
\tbl{Rank classification on $3{,}468$ held-out curves. The CNN and MLP have
matched parameter budgets and see all $1{,}229$ primes; the transformer sees
only the first $100$ ($p\le 541$), where attention's $O(T^2)$ cost is
affordable, so its accuracy is not directly comparable to the other
two.\label{tab:seq}}
{\begin{tabular}{@{}lcccc@{}}
\toprule
Model & Input & Parameters & Train acc. & Test acc. \\
\colrule
1D CNN            & $1{,}229$ primes & $39{,}714$ & $99.91\%$ & $\mathbf{99.88\%}$ \\
MLP               & $1{,}229$ primes & $39{,}922$ & $100.00\%$ & $99.13\%$ \\
Transformer       & $100$ primes  & $11{,}746$ & $99.53\%$ & $99.51\%$ \\
\botrule
\end{tabular}}
\end{table}

Table~\ref{tab:seq} reports the outcome. Both models solve the task, which is
worth stating: the rank is recoverable from $a_p$ data to nearly $100\%$
accuracy, consistent with earlier data-scientific studies of
BSD.\cite{ABH2019,Murmurations} The instructive difference is the
\emph{generalization gap}. The MLP memorizes the training set exactly
($100.00\%$) and returns $99.13\%$ on held-out curves, a gap of $0.87$ points;
the CNN, at the same parameter count, closes it to $0.03$ points ($99.91\%$
versus $99.88\%$). Weight sharing along the prime axis removes degrees of
freedom the data does not need---and what it removes is what the MLP spends on
memorization.

One caveat keeps this honest: the task is nearly saturated, so the comparison
discriminates between $99.1\%$ and $99.9\%$ rather than between success and
failure. A harder version---higher ranks, larger conductors, fewer
primes---would separate the two architectures more sharply, and the gap
reported here should be read as an indication of direction rather than of
magnitude.

\subsection{What the network learned: saliency across training}

A classifier that is right for the wrong reason is not evidence about
mathematics. Following the standard construction,\cite{Simonyan2014} define
the \textbf{per-class saliency} at prime $p$ as the average sensitivity of the
logit $S_v$ for class $v$ to the input coordinate $\ap_p$,
\begin{equation}
W_p^{(v)} \;=\; \frac{1}{|\{E:\hat v(E)=v\}|}
\sum_{\hat v(E)=v} \frac{\partial S_v(E)}{\partial \ap_p},
\label{eq:saliency}
\end{equation}
the average running over the held-out curves the network assigns to class $v$;
in JAX this is one call to \texttt{jax.grad} composed with \texttt{jax.vmap}.
Figure~\ref{fig:sal} plots the rescaled map
$\smash{\tilde W^{(v)}_p = W^{(v)}_p/\max_q|W^{(v)}_q|}$, so only the shape
matters. Evaluating Eq.~(\ref{eq:saliency}) at every epoch checkpoint turns
interpretability into a movie rather than a snapshot.

\begin{figure}[!ht]
\centerline{\includegraphics[width=\textwidth]{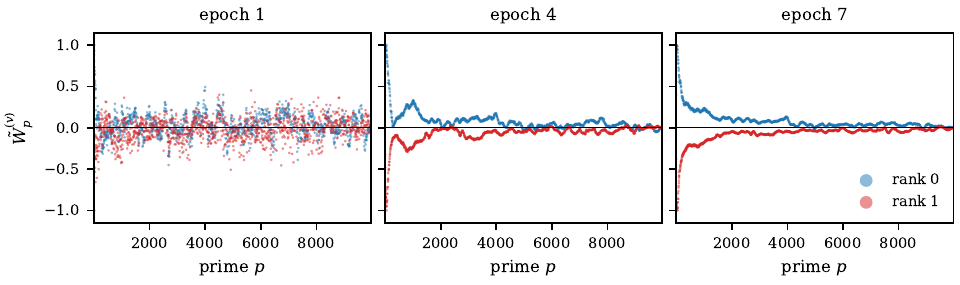}}
\caption{Rescaled per-class saliency of the 1D CNN at three training
checkpoints. At epoch~1 the map is
unstructured noise. By epoch~4 it has become smooth and antisymmetric between
the two classes, with oscillatory structure below $p\approx 4000$. By epoch~7
the sensitivity is sharply concentrated in the small primes and decays towards
zero for large $p$: the network has localized the discriminative signal in
exactly the regime where the class averages of Fig.~\ref{fig:murm} separate.}
\label{fig:sal}
\alttext{Three panels of saliency against prime at epochs one, four and seven.
The first is scattered noise; the second shows two smooth mirror-image curves
with oscillations; the third shows sharply peaked curves at small primes
decaying to zero.}
\end{figure}

Three features are worth naming. The saliency is \emph{antisymmetric} between
classes, $W^{(0)}\approx -W^{(1)}$, the correct signature for a binary
discriminant built on a signed statistic. It becomes \emph{smooth in $p$}, the
fingerprint of weight sharing and not something an MLP produces. And it
\emph{concentrates}: mass migrates over training towards $p\lesssim 2000$, the
regime where the class averages of Fig.~\ref{fig:murm}(b) separate most
cleanly. The direction of inference matters: this shows the CNN weighting the
region where the known murmuration signal is strongest, not the CNN
independently discovering murmurations.

\subsection{Attention as a different bias}

Self-attention makes no locality assumption: every position may attend to
every other, and ordering enters only through an explicit positional
encoding.\cite{Vaswani2017} The price is the $O(T^2)$ cost of the attention
matrix, so we restrict to the first $T=100$ primes ($p\le 541$) and build one
encoder block from scratch---input projection, sinusoidal positional encoding,
two-head scaled dot-product attention with residual connection and layer
normalization, position-wise feedforward, mean pooling, linear classifier.
With $11{,}746$ parameters it reaches $99.51\%$ (Table~\ref{tab:seq}).

This is not a like-for-like comparison with the CNN, which sees all $1{,}229$
primes, and we do not read it as one: the point is that a bias making no
locality assumption at all still recovers the rank from a twelfth of the
input. Attention is the more flexible bias and it is not free---quadratic cost
is what forced the truncation in the first place. It earns that cost when the
relevant interactions are long range, and deciding whether they are is a
question about the mathematics, not the model.

%============================================================================
\section{Graph Data: Quivers and Mutation Classes}
\label{sec:graph}

\subsection{Quivers, mutation, and the task}

A quiver is a directed multigraph, encoded here by a skew-symmetric exchange
matrix $B$ with $b_{ij}>0$ recording $b_{ij}$ arrows $i\to j$. The mutation
$\mu_k$ at a vertex $k$ acts by\cite{FZ2002}
\begin{equation}
b'_{ij} \;=\;
\begin{cases}
-b_{ij}, & i=k \text{ or } j=k,\\[2pt]
b_{ij} + \dfrac{|b_{ik}|\,b_{kj} + b_{ik}\,|b_{kj}|}{2}, & \text{otherwise,}
\end{cases}
\label{eq:mutation}
\end{equation}
which combinatorially means: complete every $2$-path $i\to k\to j$ by a new
arrow $i\to j$, reverse all arrows at $k$, then cancel any $2$-cycles created.
Mutation generates an equivalence relation, and the mutation class of a Dynkin
quiver is a basic invariant in the theory of cluster algebras.

We take the families $A_n$ (the path $0\to 1\to\cdots\to n{-}1$) and $D_n$ (a
path with a fork at $n{-}3$), apply between $5$ and $25$ random mutations to
each seed, deduplicate the resulting exchange matrices, and ask the model to
recover the type. After a handful of mutations the quivers carry cycles,
multiple arrows and reversed orientations and no longer resemble their Dynkin
diagrams, so this is not pattern matching. The construction follows the
setting of Ref.~\citen{QuiverGNN}, which studies mutation-class classification
with GNNs at larger scale and with explainability methods. We train on $800$
quivers with $n=7,8$ vertices and test on $100$ quivers with $n=9$---a size
never seen during training.

\subsection{DirGINE: three design decisions}

A message-passing layer updates each vertex from its neighbours,
$\bh_v' = \sigma\bigl(\sum_{u\in N(v)} f(\bh_u)\bigr)$, which in matrix form is
a single product $A\cdot f(\bH)$.\cite{Gilmer2017} Three choices turn this
skeleton into an architecture suited to quivers.

\emph{Sum, not mean.} Mean aggregation is not injective on multisets: distinct
neighbourhoods collapse to the same output. The Graph Isomorphism Network
replaces it by a raw sum followed by an MLP,
\begin{equation}
\bh_v' \;=\; \MLP\Bigl(\bh_v + \sum_{u\in N(v)} \bh_u\Bigr),
\label{eq:gin}
\end{equation}
which is provably as expressive as the $1$-Weisfeiler--Leman
test.\cite{GIN}

\emph{Direction matters.} Mutation reverses arrows, so two quivers with the
same underlying undirected graph but different orientations may lie in
different mutation classes; an orientation-blind model discards exactly the
information the task turns on. Following Ref.~\citen{DirGNN}, we split
aggregation into an incoming and an outgoing channel,
\begin{equation}
\bh_v^{(\ell+1)} \;=\; \ReLU\Bigl(W\bh_v^{(\ell)} + b
\;+\; \underbrace{\sum_{u\to v}\varphi_{\mathrm{in}}\bigl(\bh_u^{(\ell)}\bigr)}
_{\text{predecessors}}
\;+\; \underbrace{\sum_{v\to w}\varphi_{\mathrm{out}}\bigl(\bh_w^{(\ell)}\bigr)}
_{\text{successors}}\Bigr),
\label{eq:dirgine}
\end{equation}
with $\varphi_{\mathrm{in}},\varphi_{\mathrm{out}}$ two-layer MLPs. In code
this is \texttt{A\_in @ MLP\_in(H) + A\_out @ MLP\_out(H)}: still just matrix
multiplication. The resulting DirGINE\cite{QuiverGNN} uses four layers of
width $32$, a sum readout and a linear classifier ($17{,}696$ parameters), and
starts from constant vertex features---all the information is in the topology.

\emph{Locality is size-agnostic.} Nothing in Eq.~(\ref{eq:dirgine}) refers to
the number of vertices, so the same weights apply to graphs of any size.

\subsection{Equivariance for free, and generalization across size}

Because Eq.~(\ref{eq:dirgine}) is built only from adjacency products and
position-wise maps, permutation equivariance is structural rather than
learned: for any permutation matrix $P$,
\begin{equation}
\bH\bigl(PA_{\mathrm{in}}P^{\!\top},\,PA_{\mathrm{out}}P^{\!\top},\,P\bX\bigr)
\;=\; P\,\bH(A_{\mathrm{in}},A_{\mathrm{out}},\bX),
\label{eq:equivariance}
\end{equation}
and the graph-level logits, obtained by a sum readout, are exactly invariant.
The identity holds in exact arithmetic for any weights; measured numerically on
a test quiver it holds to the last bit, both errors coming out at $0.0$ rather
than at the $10^{-8}$ one would accept from an approximately equivariant model.
This is the sharpest illustration in the tutorial of what an inductive bias is:
a property holding for \emph{every} input, before training, with no data
required. A model without it would have to learn Eq.~(\ref{eq:equivariance})
from examples, and would only ever hold it approximately.

\begin{table}[!ht]
\tbl{Quiver mutation-type classification (type $A$ versus type $D$), after
eleven epochs. The model is trained on $800$ quivers with $n=7,8$ and
evaluated on $100$ quivers with $n=9$, a size never seen in training.
Independently of training, the node equivariance of
Eq.~(\ref{eq:equivariance}) and the invariance of the graph-level logits both
hold with maximum error exactly $0.0$.\label{tab:gnn}}
{\begin{tabular}{@{}lcc@{}}
\toprule
Evaluation set & Vertices $n$ & Accuracy \\
\colrule
Training set        & $7,8$ (seen)    & $100\%$ \\
Test set            & $9$  (unseen)   & $99\%$ \\
\botrule
\end{tabular}}
\end{table}

Table~\ref{tab:gnn} records the payoff. Trained only on $7$ and $8$ vertices,
the network reaches $99\%$ on quivers with $9$---a direct consequence of
locality: message passing learns a rule about neighbourhoods, and
neighbourhoods do not know how large the ambient graph is. Whether the
transfer degrades gracefully at substantially larger $n$, and whether a fixed
depth of four layers eventually becomes the binding constraint as the diameter
grows, is the natural next experiment.

%============================================================================
\section{Continuous Data: PINNs and Rigorous Verification}
\label{sec:pinn}

\subsection{The problem and the network}
\label{sec:pinn-setup}

The third inductive bias constrains the model by a differential equation
rather than a symmetry. Consider the two-point boundary value problem
\begin{equation}
-u''(x) \;=\; \pi^2\sin(\pi x),\qquad x\in[0,1],
\qquad u(0)=u(1)=0,
\label{eq:bvp}
\end{equation}
whose exact solution is $u(x)=\sin(\pi x)$. A physics-informed neural
network\cite{PINN} replaces the discretization of Eq.~(\ref{eq:bvp}) by a
network $u_\theta$ trained to make the residual small,
\begin{equation}
\mathcal{L}(\theta) \;=\;
\frac{1}{N}\sum_{i=1}^{N}\bigl|-u_\theta''(x_i) - f(x_i)\bigr|^2
\;+\; \lambda\bigl(|u_\theta(0)|^2 + |u_\theta(1)|^2\bigr),
\qquad f(x)=\pi^2\sin(\pi x).
\label{eq:pinnloss}
\end{equation}
The derivative $u_\theta''$ is not a finite difference but the exact
derivative of the network, obtained by composing \texttt{jax.grad} with itself,
so Eq.~(\ref{eq:pinnloss}) carries no discretization error at the collocation
points. We take $u_\theta$ to be a $\tanh$ network of shape $1\to 32\to 1$
($97$ parameters), $N=50$ collocation points uniformly spaced in
$[0.01,0.99]$, $\lambda=10$, and $10^4$ Adam steps at learning rate $10^{-3}$.
The network is deliberately shallow: the over-approximation (``wrapping'')
incurred by the interval analysis of Sec.~\ref{sec:interval} compounds with
depth, so verifiability is itself an architectural constraint.

Training converges to $\mathcal{L}=3.7\times 10^{-4}$, and $u_\theta$ agrees
with $\sin(\pi x)$ to a maximum pointwise deviation of $9.0\times10^{-4}$ on a
fine grid (Fig.~\ref{fig:pinn}(a),(b)). This is where most PINN
demonstrations stop---and, for a problem whose exact solution is unknown, where
the honest questions begin. The residual was checked at $50$ points. What
happens on the rest of the continuum?

\subsection{Interval arithmetic: from a good fit to a proof}
\label{sec:interval}

An interval $[a,b]$ denotes $\{x\in\RR: a\le x\le b\}$, and interval arithmetic
propagates guaranteed enclosures through
computation:\cite{Moore2009,Tucker2011}
\begin{equation}
[a,b]+[c,d] = [a+c,\,b+d],
\qquad
[a,b]\cdot[c,d] = [\min S,\,\max S],
\quad S=\{ac,ad,bc,bd\}.
\label{eq:interval}
\end{equation}
Two points make the guarantee real rather than nominal. First, even $a+c$ is
not computed exactly---IEEE 754 rounds to the nearest representable
double\cite{IEEE754}---so every lower bound is nudged down and every upper
bound up by one unit in the last place (\texttt{np.nextafter}), and a
length-$k$ inner product is widened by $k$ ULPs. Second, interval arithmetic
ignores correlations between operands: $x-x$ over $[0,1]$ returns $[-1,1]$,
not $\{0\}$. Enclosures are thus always valid and generally pessimistic, and
controlling that pessimism is the engineering problem.

To bound $u_\theta''$ over a whole subinterval we propagate the jet
$(u,u',u'')$ through the network in interval form---forward-mode automatic
differentiation by hand, every intermediate quantity an enclosure. A linear
layer $z=Wu+b$ gives $z'=Wu'$, $z''=Wu''$; for $h=\tanh(z)$, writing
$s=\operatorname{sech}^2(z)=1-\tanh^2(z)$,
\begin{equation}
h' \;=\; s\,z',
\qquad
h'' \;=\; -2\tanh(z)\,s\,(z')^2 + s\,z''.
\label{eq:jet}
\end{equation}
Each operation in Eq.~(\ref{eq:jet}) admits a rigorous enclosure: $\tanh$ is
monotone, so its enclosure is exact up to rounding; squaring is tightened by
hand when the interval straddles zero; $\sin$ is enclosed by its endpoint
values, corrected at any critical point $\pi/2+k\pi$ inside the interval. The
full scheme is Algorithm~\ref{alg:verify}, closely related to interval bound
propagation as used for neural network verification.\cite{IBP}

\begin{algorithm}[!ht]
\caption{Rigorous residual bound for a trained PINN}
\label{alg:verify}
\begin{algorithmic}[1]
\Require trained parameters $\theta$; domain $[0,1]$; subdivision count $M$
\Ensure $\rho$ with $|-u_\theta''(x)-f(x)|\le\rho$ for \emph{every}
$x\in[0,1]$
\State partition $[0,1]$ into $M$ intervals $I_m$ of width $1/M$
\For{each $I_m$ (vectorized over $m$)}
  \State initialize the jet $(u,u',u'')\gets(I_m,[1,1],[0,0])$
  \For{each layer}
    \State propagate $(u,u',u'')$ by Eq.~(\ref{eq:jet}) in interval
    arithmetic with outward rounding
  \EndFor
  \State $R_m \gets (-u'') - \pi^2\cdot\textsc{IntervalSin}(\pi I_m)$
\EndFor
\State \Return $\rho \gets \max_m \max\bigl(|\underline{R_m}|,
|\overline{R_m}|\bigr)$
\end{algorithmic}
\end{algorithm}

\begin{figure}[!ht]
\centerline{\includegraphics[width=\textwidth]{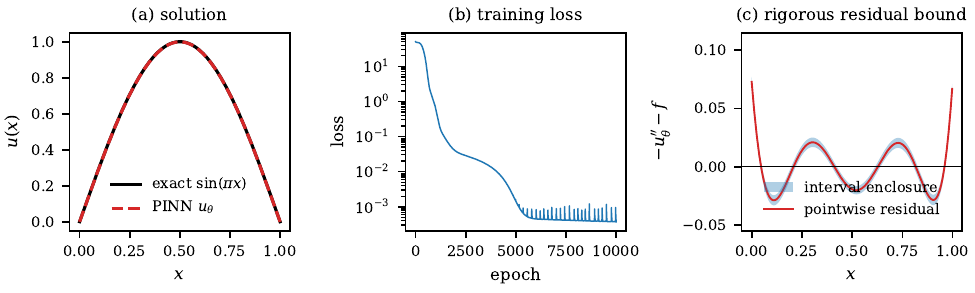}}
\caption{(a) The trained PINN against the exact solution $\sin(\pi x)$.
(b) Training loss. (c) The interval enclosure of the residual
$-u_\theta''-f$ over $10^4$ subintervals (band) against the pointwise residual
at subinterval centres (line). The enclosure is tight; the residual oscillates
within about $3\times10^{-2}$ across the interior and rises steeply at both
endpoints.}
\label{fig:pinn}
\alttext{Three panels. The first shows the PINN and exact solution
indistinguishable. The second shows loss decreasing over four orders of
magnitude. The third shows a residual oscillating at the level of a few
hundredths across the interior and rising steeply to about seven hundredths at
both endpoints, with a narrow enclosure band around it.}
\end{figure}

\subsection{What verification reveals}

Running Algorithm~\ref{alg:verify} with $M=10^4$ certifies
\begin{equation}
\bigl|-u_\theta''(x) - f(x)\bigr| \;\le\; 7.70\times10^{-2}
\qquad\text{for every } x\in[0,1],
\label{eq:resbound}
\end{equation}
a statement about every point of the domain rather than about a finite sample.
The enclosure is not loose: Fig.~\ref{fig:pinn}(c) shows it hugging the
pointwise residual evaluated at subinterval centres, so at $M=10^4$ the
wrapping effect has not yet swamped the estimate.

Two things follow. The first is a matter of magnitude. The converged loss
$\mathcal{L}=3.7\times10^{-4}$ is dominated by its residual term, so the
residual at the $50$ collocation points has root-mean-square about
$1.9\times10^{-2}$; the
\emph{guaranteed} worst case over the continuum, Eq.~(\ref{eq:resbound}), is
four times larger. A training curve that has flattened at $10^{-4}$ is
reporting an average over the points it was given, and averages over samples
are not bounds.

The second is a matter of shape. The residual is far from uniform
(Fig.~\ref{fig:pinn}(c)): it oscillates between about $\pm3\times10^{-2}$
across the interior, through five local extrema, and then rises steeply within
the outermost few per cent at each end, attaining its maximum at $x=0$ and
$x=1$. A sufficiently dense pointwise scan would suggest the same picture; what
the interval computation adds is the guarantee that the bound holds
\emph{between} the sample points too, so the maximum cannot be hiding in a gap
the scan stepped over. This is the practical argument for verification over
validation: sampling tells you about the samples.

Every quantity entering Eq.~(\ref{eq:resbound}) is an interval computation
with outward rounding. Granted that the library's $\tanh$ and $\sin$ are
accurate to within the one unit in the last place by which each enclosure is
widened, the inequality is therefore a statement about the trained network
rather than a statistic---and, unlike the $9.0\times10^{-4}$ deviation of
Sec.~\ref{sec:pinn-setup}, it is obtained without any knowledge of the exact
solution. Converting a residual bound of this kind into a bound on
$\|u-u_\theta\|$, and sharpening the constants involved, is a subject in its
own right.\cite{MishraMolinaro}

%============================================================================
\section{Discussion}
\label{sec:outlook}

Three data types, three inductive biases, three kinds of payoff. For the
sequential data of Sec.~\ref{sec:seq}, matching architecture to structure
bought \emph{generalization}: at identical parameter budget, weight sharing
cut the train--test gap from $0.87$ points to $0.03$, and saliency confirmed
the network weighting the primes where the murmuration signal lives. For the
graph data of Sec.~\ref{sec:graph}, it bought an \emph{exact guarantee}:
permutation equivariance held to machine zero before a single gradient step,
and locality carried a model trained on $7$ and $8$ vertices to $99\%$ on $9$.
For the continuous data of Sec.~\ref{sec:pinn}, it bought
\emph{verifiability}: because the model is a small smooth function of one
variable rather than a lookup table, interval arithmetic bounded its residual
at every point of the domain---and that guaranteed bound turned out to be four
times the residual the training loss was reporting.

These are not three tricks but one habit. Before choosing a model, ask what
the data is invariant under, what it is indexed by, and what equations it
satisfies; then pick the architecture that already respects those answers. In
mathematics the answers are unusually well specified---the symmetry group, the
differential equation, the irrelevance of vertex labels---and that specificity
is an advantage the field should spend rather than ignore.

Several directions follow. The rank task should be pushed where architecture
separates models more sharply: higher ranks, larger conductors, or restriction
to the small primes where the signal concentrates. For the mutation classifier,
the obvious tests are more Dynkin types, larger quivers, and explainability
methods in the spirit of Ref.~\citen{QuiverGNN}. The verification pipeline of
Sec.~\ref{sec:pinn} has the broadest reach: rigorous a posteriori certification
of learned objects---solutions of differential equations, but equally metrics,
forms and other geometric data produced by networks---is where machine learning
and mathematics can meet without either giving up its standards of proof.

The three tutorial notebooks (\texttt{block1\_sequential.ipynb},
\texttt{block2\_graph.ipynb}, \texttt{block3\_pinn.ipynb}), together with the
elliptic curve dataset and Colab links, are available at
\url{https://github.com/yidiq7/Banff-Tutorial-Architecture}. Each is
self-contained, depends only on JAX, NumPy and Matplotlib, and runs on CPU in
minutes.

\section*{Acknowledgments}
The author thanks the DANGER organizers---Edward Hirst, Elli Heyes, Miranda
Cheng and Sergei Gukov---for the invitation and for assembling such a
stimulating meeting, and the tutorial participants for questions that
sharpened several of the points above.
% TODO: add grant / IAIFI support acknowledgment, and thank collaborators

\section*{ORCID}
\noindent Yidi Qi -- \url{https://orcid.org/0000-0002-2094-0257}

\bibliographystyle{ws-ijdsms}
\bibliography{references}

% <<< END PASTE: TUTORIAL 2 DOCUMENT BODY <<<

\end{tutorial}

%=========================================================================
% TUTORIAL 3 -- Jessica N. Howard
% Source: third_tutorial_Jessica/ThirdTutorial.tex
% The document body from this source is embedded unchanged below.
% This tutorial retains its own appendix and reference list.
% Its bibliography command points to the shared references.bib database.
%=========================================================================
\begin{tutorial}{3}{Jessica N. Howard}
  {Tips and tricks for analyzing and presenting machine learning results}
  {.}{references}

% >>> BEGIN PASTE: TUTORIAL 3 DOCUMENT BODY (UNCHANGED) >>>

% \markboth{Authors' Names}{Instructions for Typing Manuscripts (Paper's Title)}

\catchline{}{}{}{}{}
%
%%%%%%%%%%%%%%%%%%%%%%%%%%%%%%%%%%%%%%%%%%%%%%%%%%%%%%%%%%%%%%%%%%%%

\title{Techniques Session 3:\\ Tips and tricks for analyzing and presenting machine learning results}

\author{Jessica N. Howard}

\address{University of California, Santa Barbara, USA\\
Kavli Institute for Theoretical Physics, Santa Barbara, California, USA\\
jnhoward@kitp.ucsb.edu}

\maketitle

% \begin{history}
% \received{(Day Month Year)}
% \revised{(Day Month Year)}
% \accepted{(Day Month Year)}
% \published{(Day Month Year)}
% \end{history}

\begin{abstract}
These lecture notes summarize the main points from the third tutorial session of the workshop ``\emph{DANGER: Data, Numbers, and Geometry}'' held in April 2026 at the Banff International Research Station for Mathematical Innovation and Discovery (BIRS).
The accompanying code and slide presentation can be found at 
\faGithub~\href{https://github.com/jnhoward/Banff-Tutorial/}{jnhoward/Banff-Tutorial/}.
\end{abstract}

% \keywords{Keyword1; keyword2; keyword3.}

% \ccode{Mathematics Subject Classification 2020: 22E46, 53C35, 57S20}

\section{Introduction}

At its core, Machine Learning (ML) is an incredibly powerful tool because of its ability to extract high-dimensional, non-linear relationships between features in data.%
\footnote{When one hears ``data'' they may think of standard ML datasets (MNIST~\cite{MNIST}, CIFAR10~\cite{CIFAR10}, etc.) or observations of an experiment. Here, we take the broader definition of data as being a numerical representation of the problem you are trying to solve. Thus, one can \emph{define} data that is relevant to the theoretical or mathematical problem at hand.}
But the high dimensionality and probabilistic nature of data pose two key challenges:
\begin{enumerate}
    \item \textbf{High $\rightarrow$ Low:}
    Ultimately, we need to compress our machine-learned relationships into much lower dimensional spaces. 
    This is partly because we can only really think intuitively in low dimensions (2, 3, maybe 4), making this dimensionality reduction useful for interpretation.
    More pragmatically, we also have to represent these machine-learned relationships in two-dimensional plots that we can use in papers.
    Therefore, we typically rely on summary metrics to understand our high-dimensional, machine-learned results.
    \item \textbf{Uncertainties and generalizability:} If an ML model trains on certain random samples of data, will the learned relationships still work on new random samples? What are the statistical uncertainties we can expect on the summary metrics? 
\end{enumerate}

\noindent The goal of these notes is to provide a brief survey of common techniques and best practices to approach these challenges.
The target audience are individuals who are just getting started with ML.
Thus, we limit our scope to introductory concepts, but will include references where one can learn more.
The reader should be familiar with basic ML terminology such as loss function and training/validation/testing data split.

\section{Summary Metrics for Common ML Tasks}

A trained ML model typically defines a high-dimensional relationship that cannot be inspected directly.
Summary metrics are low-dimensional representations of model performance.
Note that the loss function used to train your ML model is one such low-dimensional representation. 
However, it is generally advantageous to look at other metrics in addition to the loss function to fully assess performance.

The appropriate metric(s) depends strongly on the type of ML task.
We can generally split ML tasks into three categories: (1) Generative modeling, (2) Regression, (3) Classification. 
We will briefly cover common summary metrics for (1) and (2) and then do a much deeper dive on (3) in Section~\ref{sec:Classification} since it is the most theoretically well-understood case.

\subsection{Generative Modeling Task}\label{sec:GenModeling}

For a generative modeling task, the objective is to learn a model distribution $p_{\rm model}(x)$ that approximates the target distribution $p_{\rm target}(x)$. 
For example, a model that generates images of cats is implicitly learning the distribution of all possible cat images. 
Evaluating such a model therefore requires a summary metric that quantifies the difference between the model and target distributions.

It turns out there are not many ways to robustly compare two high-dimensional probability distributions.
One can use distance metrics such as Maximum Mean Discrepancy (MMD)~\cite{JMLR:v13:gretton12a}, which is an integral probability metric that measures the largest difference between expectations over kernel functions within a chosen class, or Optimal Transport distances~\cite{flamary2021pot}, which quantify the cost of transforming $p_{\rm model}$ into $p_{\rm target}$.
Another approach is to train a classifier to distinguish model-generated samples from target samples~\cite{Plehn:2022ftl}. 
This works because an optimal classifier is closely related to the likelihood ratio between the two distributions. 
If the distributions have overlapping support and their probability densities can be estimated, one may also use f-divergences such as the Kullback-Leibler (KL) divergence~\cite{Goodfellow-et-al-2016}. 
However, reliable density estimation may require many samples, particularly in high dimensions. 

Given the limited number of options, it is common for one of these metrics to already appear in the model’s training loss. 
Thus, ideally, model performance should be evaluated using another metric as well.

\subsection{Regression Task}\label{sec:Regression}

For a regression task, the objective is to learn a function, $f_{{\rm model}}(x)$, that approximates the target function, $f_{{\rm target}}(x)$, across the domain of $x$. 
For continuous outputs, such as predicting the age of a cat from an image, a common summary metric is the Mean Squared Error (MSE),
\begin{equation}
    \mathrm{MSE}=\frac{1}{n}\sum_i\left[f_{\mathrm{model}}(x_i)-f_{\mathrm{target}}(x_i)\right]^2,
\end{equation}
which measures the average squared difference between the predicted and target values. 
When the target values have known uncertainties, $\sigma_i$, these differences can instead be weighted to calculate the chi-squared statistic,
\begin{equation}
    \chi^2=\sum_i\frac{\left[f_{\mathrm{model}}(x_i)-f_{\mathrm{target}}(x_i)\right]^2}{\sigma_i^2}~.
\end{equation}
If every $\sigma_i=1$, this becomes the residual sum of squares.

Categorical outputs, such as the eye color of the cat in the image, can be represented numerically using one-hot encoding and compared using similar metrics (e.g. MSE). 
Namely, given an image, $x$, say the true color of the cat's eyes is ``Teal''. We can turn this label into a continuous vector,
\begin{equation}
    \text{Teal} \quad \rightarrow \quad  (1,0,\ldots,0) \in \mathbb{R}^{N_{\rm classes}}
\end{equation}
where $N_{\rm classes}$ is the number of categories, which in this case is the number of possible cat eye colors.
In practice, however, categorical labels are usually best treated in a multi-class classification setup.

\subsubsection{Regression vs Classification}\label{sec:RegressionVsClassification}

When the task is categorical in nature, regression and classification become very similar.
And in fact, one could learn to classify data by regressing the encoded class label vectors.
In this case, the model would output a continuous vector which is then matched to the appropriate unit target vector via an MSE loss.

On the other hand, a classification model's output is typically interpreted as a categorical distribution across the model classes.
In this case, it is advantageous to change our loss function to categorical cross-entropy (i.e. cross entropy evaluated on these categorical distributions).%
\footnote{Note that the built-in cross entropy loss functions within ML libraries typically assume that the output vector is the unnormalized logits for each class.
These logits need not be positive nor sum to 1. 
This is then converted into the categorical probabilities for each class within the loss function calculation. 
So there is a question of whether you count this step as part of the model or a deterministic operation in the loss.
Thus, as we will see in the next section, it is always possible to turn a continuous score into a class label which blurs the line between regression and classification further.}
For example, when trying to match the correct label vector $(1,0,\ldots,0)$ a regression algorithm may predict $(1,0.1, 0,\ldots,0)$ while a classification algorithm may predict $(0.9,0.1,0,\ldots,0)$ which is appropriately normalized.
Of course, one could also use MSE on this normalized vector, but cross-entropy turns out to work better in practice. 
Using MSE to train a classifier in this case will result in slower convergence during training due to vanishing gradients~\cite{Goodfellow-et-al-2016, BOSMAN2020113, Golik2013}.

\section{Binary Classification Analysis: From Scores to ROC Curves}\label{sec:Classification}

Our task will be to assign one of two labels to each data point $x$. We will call the two classes \background~and \signal.
Let $C(x)$ be a binary classifier which assigns an example data point, $x_i$, a one-dimensional score, $C(x_i) = s_i$. 
Note that this score need not be a strict probability~\cite{FAWCETT2006861}. 
However, we will interpret higher (lower) scores as indicating that $x_i$ belongs to class \signal~(class \background). 
But how large is large enough to be considered as belonging to class \signal?
Ultimately, we need to turn this continuous score into a binary decision about the identity of $x_i$. 
We can do this by imposing a threshold defining what is ``\signal~enough''.
Changing this threshold will produce different labels for the same $x_i$. 
The goal is to optimize this choice so that the predicted labels are as close as possible to the true ones.

\subsection{Score Thresholds and the Confusion Matrix}
After choosing a threshold, the data points, $x_i$, with scores, $s_i$, above the threshold, $T$, are classified as \signal~(``accepted''), while those below it are classified as \background~(``rejected''). 
The True Positive Rate (TPR) is the fraction of actual \signal~data points above the threshold, and the False Negative Rate (FNR) is the fraction of actual \signal~data points below it, so that $\mathrm{TPR}+\mathrm{FNR}=1$. 
Similarly, the False Positive Rate (FPR) is the fraction of actual \background~ data points above the threshold, while the True Negative Rate (TNR) is the fraction below it, with $\mathrm{FPR}+\mathrm{TNR}=1$.
This naming comes from medical fields where the goal is usually to test for a particular disease. 
Thus, the two classes are ``Positive'' and ``Negative'' instead of \signal~and \background.

These four quantities can be used to form the confusion matrix,
\begin{equation}
    \begin{pmatrix}
        \text{TP} & \text{FN}\\
        \text{FP} & \text{TN}
    \end{pmatrix}
    =
    \begin{pmatrix}
        \text{TPR} ~N_{\signal} & \text{FNR} ~N_{\signal}\\
        \text{FPR} ~N_{\background} & \text{TNR}~N_{\background}
    \end{pmatrix}
\end{equation}
where $N_{\signal}$ ($N_{\background}$) is the number of \signal~(\background) data points in the sample.
The confusion matrix summarizes the classifier's behavior at a particular threshold. 
Several named metrics can be derived from this such as accuracy$= \frac{\text{TP} + \text{TN}}{N_{\signal} + N_{\background}}$ and F1-score$= \frac{2 \text{TP}}{2 \text{TP} + \text{FP} + \text{FN}}$.

In high-energy particle physics terminology, the classes are \signal~= signal and \background~= background. Thus, the TPR is also called the signal efficiency, $\epsilon_S$, since it is the amount of signal accepted. Similarly, the FPR is the background efficiency, $\epsilon_B$, and the TNR is the background rejection rate, $1-\epsilon_B$.

\subsection{The Receiver Operating Characteristic (ROC) Curve}

\begin{figure}[ht]
    \centering
    \includegraphics[width=\linewidth]{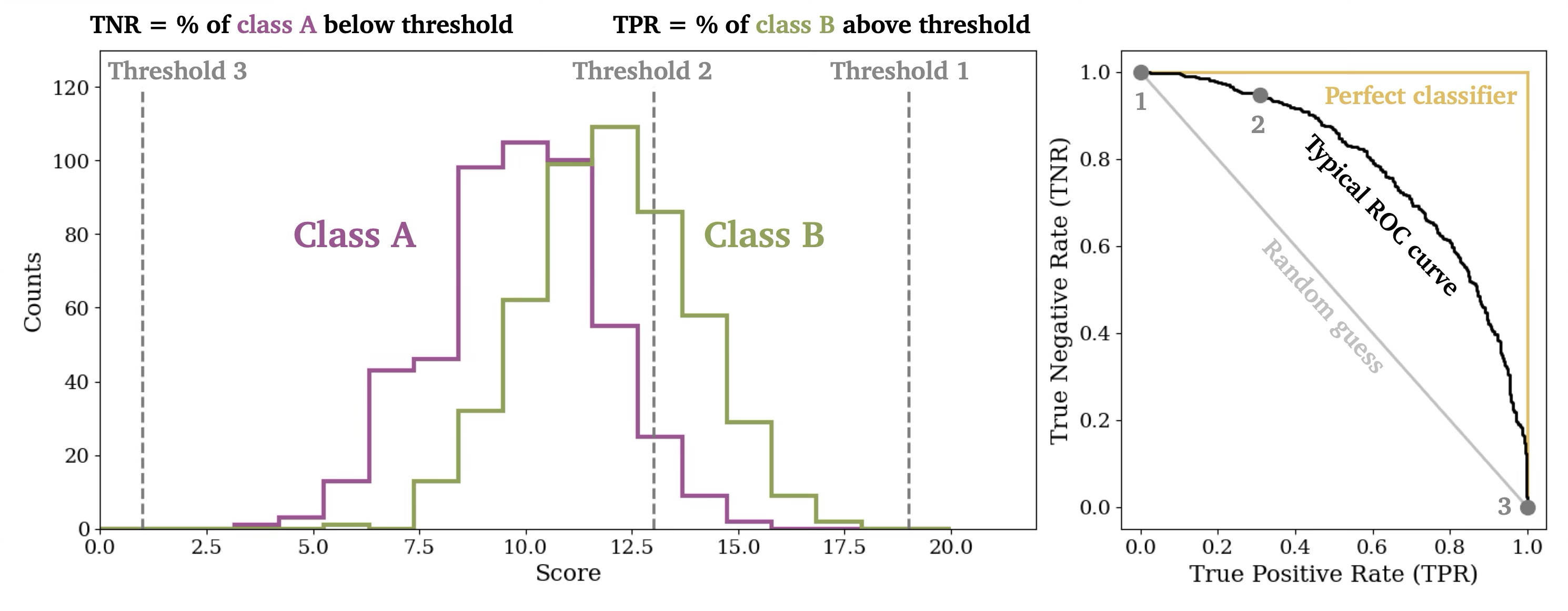}
    \caption{\textbf{Figure showing how a ROC curve can be constructed from a score distribution.} Various threshold values are swept from right to left. For each threshold the TPR (fraction of class \signal~scores falling above the threshold) and TNR (fraction of class \background~scores falling below the threshold) are measured. 
    These (TPR, TNR) points are collected to form the ROC curve. 
    The points on the ROC curve corresponding to the three thresholds are indicated.
    }
    \label{fig:ROC_curve}
\end{figure}

Varying the score threshold changes the tradeoff between TPR and TNR. 
Consider the example in Fig.~\ref{fig:ROC_curve} which shows three possible thresholds.
Threshold 1 rejects all data points, giving $(\mathrm{TPR},~\mathrm{TNR}) = (0, 1)$.
At the opposite extreme, threshold 3 accepts all data points, giving $(\mathrm{TPR},~\mathrm{TNR}) = (1, 0)$. 
Threshold 2 provides a more useful compromise, accepting $0.31$ of the \signal~data points while rejecting $0.946$ of the \background~data points. 
Continuously moving the threshold from right to left traces out the receiver operating characteristic (ROC) curve, which summarizes the available tradeoffs between signal acceptance and background rejection.

A common single-number summary of the ROC curve is the area under the curve (AUC). 
An ideal classifier has $\text{AUC}=1$, while random guessing gives an $\text{AUC}=0.5$. 
While useful, the AUC can have some drawbacks since it is compressing the entire ROC curve into a single number.
Importantly, it does not indicate which threshold should be used. 
Two classifiers might have the same AUC but one may be more performant for a particular threshold of interest (for example very high TNR).
Ideally, one can compare ROC curves directly, however when testing many models at once this can quickly become cumbersome.
So AUC is used as a standard proxy for performance, however it can be useful to look at additional ROC-derived metrics.

\subsubsection{Conventions and Alternate Visualizations}
Typically, you will see ROC curves displayed as either FPR on the horizontal axis and TPR on the vertical axis or TPR on the horizontal axis and TNR on the vertical axis. 
The latter is particularly common in experimental high-energy particle physics.
However, the majority of state-of-the-art ML papers tend to plot an alternate ROC curve: TPR on the horizontal axis and 1/FPR on the vertical axis. 
This alternative display is particularly useful when comparing classifiers which have started to saturate performance (AUC $\gtrsim 0.90$) since it will accentuate small, yet significant, differences.

\subsection{Extending to Multiple Classes}

Multi-class classification extends binary classification to problems with more than two possible category labels. 
The primary change occurs in the loss function
where we now use categorical cross-entropy~\cite{Goodfellow-et-al-2016},
\begin{equation}
    \mathcal{L}_{\rm CE} = -\sum_{c=1}^{N_{\rm classes}}q_c\log(p_c)\,
\end{equation}
where $N_{\rm classes}$ is the number of classes, $q_c$ is the true label (i.e. $1$ or $0$), and $p_c \in (0,1]$ is the predicted probability for class $c$.
For example, say we have $N_{\rm classes}=3$. For a data point $x_i$ we have $(q_1, q_2, q_3) = (1,0,0)$, meaning it belongs to class 1. 
And say our classifier predicts $(p_1, p_2, p_3) = (0.9,0.09,0.01)$. 
Then the categorical cross-entropy is $- \log (0.9) = 0.1054$.
When $N_{\rm classes}=2$ categorical cross-entropy simply reduces to the binary case.
For visualizing the results, performance is often analyzed through pairwise comparisons. 
For each matchup we can then construct ROC curves as in the binary case.

\section{Statistical Uncertainties and Dealing with Limited Data}

The metrics that we calculate inherit randomness from the data we use to calculate them. 
So we can ask questions like, ``how likely was it that we got a particular $\text{AUC}$ value?'' and ``what happens if we used a larger data sample?''.
In other words, what are the statistical properties of our metrics?

\subsection{Metrics as Random Variables}

Any performance metric $M$ will be estimated from a finite test set, $X=\{x_1,\ldots,x_n\} \rightarrow \hat{M}(X)$. 
Since $X$ is a random variable, $\hat{M}(X)$ will be as well.
Thus, a different sample, $X_j$, drawn from the same underlying distribution will generally produce a different value of $\hat{M}_j = \hat{M}(X_j)$.
This means the estimate $\hat{M} \approx M$ will have a statistical uncertainty coming from the randomness in $X$.

We can imagine calculating $\hat{M}_j$ for many different $X_j$ samples and forming a histogram of the values. 
From this we can then ask about the statistical properties of the distribution of $\hat{M}$ values. 
For example, what is the average and standard deviation of this distribution?
How likely is a particular value?
Ultimately, this will help us infer what the true value of $M$ is most likely to be.

This process is best illustrated via a classic example from statistics: estimating the standard error on the mean.
In this case, our metric $M$ is the true mean, $\mu$, of a distribution, $p(z)$.
From a random sample $Z_j = \{ z_1, \ldots, z_n \mid z_i \sim p(z)\}$, we can get an estimate of the mean as $\hat{\mu}_j = \frac{1}{n}\sum_{i=1}^n z_i$.
We can repeat this for different $Z_j$ to get a distribution of $\hat{\mu}$ values.
Alternatively, we could think of this as sampling some distribution $q(\hat{\mu})$.
For sufficiently large $n$ we know that $q(\hat{\mu})$ will be approximately normally distributed for any well-behaved $p(z)$ by the Central Limit Theorem (CLT). 
Let $\mu_0$ ($\sigma_0^2$) be the true mean (variance) of $q(\hat{\mu})$. 
Then the original mean of $p(z)$ can be estimated as $\mu = \mu_0 \pm \sigma_0$. 
Here, $\sigma_0$ is known as the standard error on the mean.
The same procedure works for other metrics. In general, let $\mu_0, \sigma_0^2$ be the true mean and variance of the distribution of estimates, $q(\hat{M})$, of a metric $M$. 
Then we say that $M = \mu_0 \pm \sigma_0$.

We can estimate $\mu_0, \sigma^2_0$ in practice by constructing an empirical estimate of $q(\hat{M})$.
When sufficient independent data are available, $\mu_0, \sigma_0^2$ can be estimated by dividing the data into $N$ independent test sets, $X_1,\ldots, X_N$, each containing $n$ data points, and evaluating the metric on each set. 
The resulting ensemble $\{\hat{M}_1,\ldots,\hat{M}_N\}$ can then be used to estimate both the expected metric value and its variance,
\begin{equation}
    \hat{\mu}_0 = \frac{1}{N} \sum_j \hat{M}_j \quad \text{ and } \quad \hat{\sigma}_0^2 = \frac{1}{N} \sum_j \left(\hat{M}_j - \hat{\mu}_0\right)^2~.
\end{equation}
such that now $M \approx \hat{\mu}_0 \pm \hat{\sigma}_0$.

This procedure is preferable when feasible, but it requires a total of $N n$ independent samples.
In practice, one typically would like at least $N \gtrsim 5$ and $n\gtrsim 30$.
The latter is a familiar rule-of-thumb for when $q(\hat{M})$ will be approximately Gaussian by the CLT, however it depends heavily on the test statistic and assumes that the distribution is reasonably well-behaved. 
Practically, $n$ will usually be far larger. 
And, in general, the larger $n$ and $N$ are, the better since substantially larger test sets  provide more reliable  estimates of $\mu_0$ and $\sigma_0^2$.
Concretely, in ML applications one can expect $n = 10^3 - 10^5$ and $N = 5 - 10$.
In cases where there is a limited amount of independent data, one must employ resampling methods to construct $q(\hat{M})$.

\subsection{Resampling Methods}

When independent test data are limited, resampling methods provide an alternative way to construct an ensemble of test sets $X_1,\ldots,X_N$ from a single sample $X={x_1,\ldots,x_n}$. 
Note that these methods still assume that $n$ is sufficiently large for the empirical distribution of $X$ to approximate the underlying distribution $p(x)$ reasonably well. 
Moreover, $X$ must contain held-out test data only;  none of its observations should have been used to train the model or select its architecture and hyperparameters during validation.
Two common resampling methods are the bootstrap and the jackknife~\cite{doi:10.1137/1.9781611970319}. 

In bootstrap resampling, each new test set $X_j$ is constructed by drawing $n$ observations from $X$ \emph{with replacement}. 
Consequently, some observations may appear multiple times in a bootstrap sample, while others may not appear at all. 
The metric is evaluated on these bootstrapped samples, and the resulting values $\{\hat M_1,\ldots,\hat M_N\}$ are used to estimate $\mu_0$ and $\sigma_0^2$.

In Jackknife resampling, we instead construct $n$ test sets of size $n-1$. 
Each set is formed by leaving out a different observation from $X$, so that every observation is omitted exactly once.
For example $X_1 = \{x_2, \ldots, x_n\} \rightarrow \hat{M}_1$.
Therefore, we evaluate the metric on $N=n$ sets of samples. 
The Jackknife estimates are then
\begin{equation}
    \begin{aligned}
        \bar{M} =& \frac{1}{n} \sum_j \hat{M}_j\\
        \sigma^2_{\rm jack} =& \frac{n-1}{n} \sum_j \left(\hat{M}_j - \bar{M}\right)^2\\
        b_{\rm jack} =& (n-1) \left(\bar{M} - \hat{M}\right)~.
    \end{aligned}
\end{equation}
where $\hat{M}$ is the metric evaluated on all $n$ samples (i.e. $X \rightarrow \hat{M}$) and $b_{\rm jack}$ is the Jackknife estimated bias. 
Then, $\hat{\mu}_0 = \hat{M} - b_{\rm jack}$ and $\hat{\sigma}_0^2 = \sigma^2_{\rm jack}$ ~\cite{doi:10.1137/1.9781611970319}.

The exercise in Section~\ref{sec:Ex_estimateUncertainty} compares both of these resampling methods to the method of splitting the test data into independent sets.
For details on the tradeoff between bootstrap and jackknife resampling see Refs.~\cite{doi:10.1137/1.9781611970319, e89fac9c-03d7-3e22-aa30-08f5596f8fce}.

\section{Plot Design: Effectively Communicate Scientific Results}

Plots are an incredibly efficient and effective way to communicate your scientific results. 
Thus, it is worth taking the time to carefully design a paper's plots.
Doing so will ensure that the impact of your science is clear.

This section covers what questions you should ask yourself in choosing how to present your results. 
In ML, this is particularly challenging because we are significantly reducing the dimensionality of our solution and are often comparing many models and metrics at once.

\subsection{Questions to ask yourself}

Every plot should tell a story, and the goal is to identify the most concise and compelling way to tell it. 
When designing a plot, ask what the most important take-away messages are and which relationships among the results should be emphasized. 
It is also useful to distinguish between information that must be shown graphically and information that can be stated more efficiently in the text. 
Because figures may be the first (and sometimes only) part of a paper that a reader examines closely, this process deserves considerable time and care.

The format of the paper also imposes practical constraints. 
For example, whether the paper uses a single- or double-column layout determines which figure dimensions will remain legible. 
The appropriate type of visualization should likewise be chosen based on the scientific message. 
A standard two-dimensional plot is often the clearest option, while a color map can encode a meaningful third variable. 
Fully three-dimensional plots can also work, however the final visual representation depends significantly upon the 2D projection that is taken. 
Thus, as a rule of thumb, three-dimensional plots should be used only when the added dimension is essential to understanding the result.
Additionally, care should be taken in selecting the projection plane that most effectively communicates the relationship.

\begin{figure}
    \centering
    \includegraphics[width=0.9\linewidth]{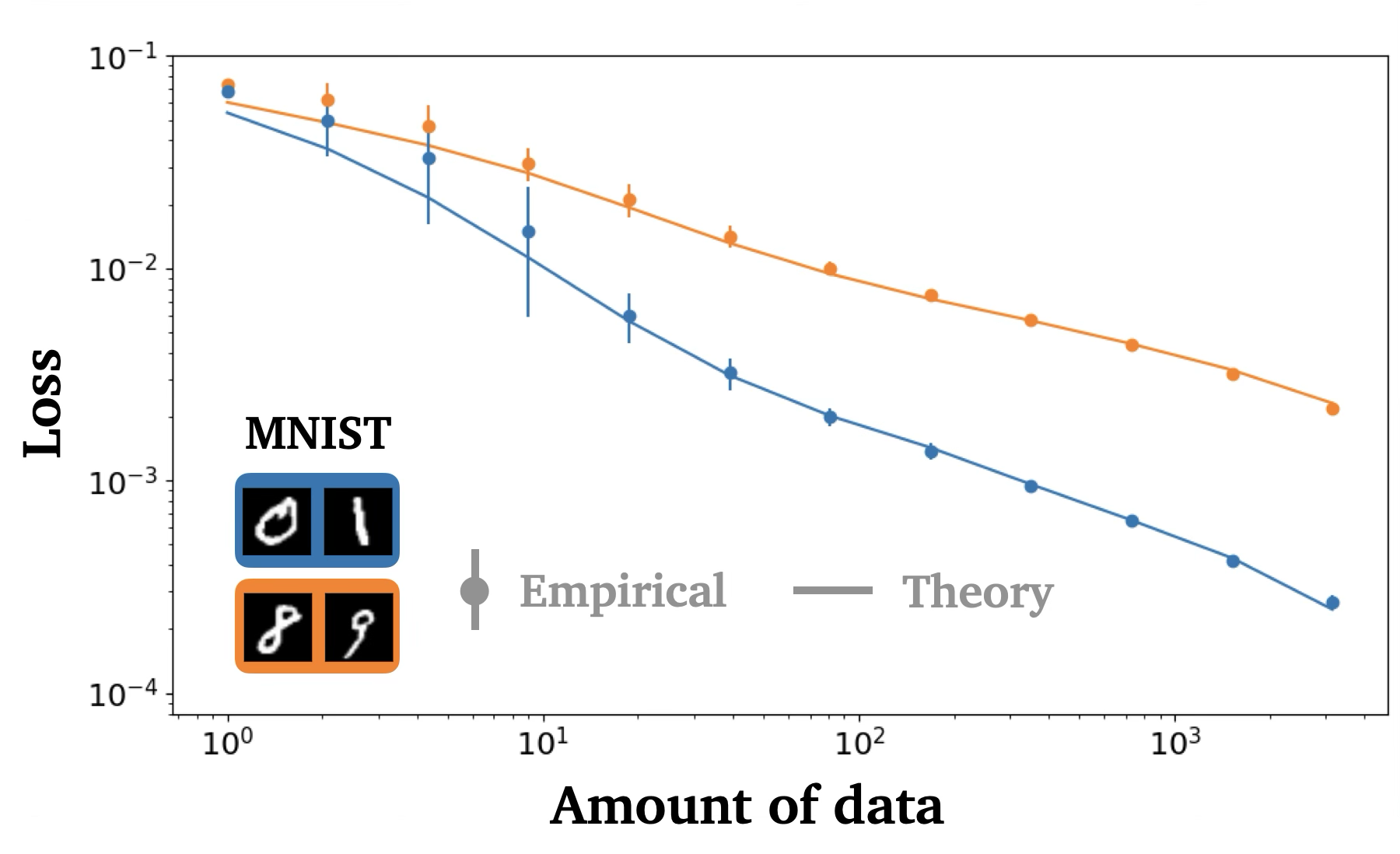}
    \caption{\textbf{Example of how a plot can be used to efficiently tell a story.}     We have modified a figure from Ref.~\protect\cite{Howard:2024bbi} to convey three key points about the result: (1) theoretical predictions agree closely with the empirical measurements, (2) the losses exhibit semi power-law behavior, (3) the two classification tasks follow quantitatively different trends.}
    \label{fig:storyOfThePlot}
\end{figure}

As an example, consider the learning curves in Fig.~\ref{fig:storyOfThePlot}, which show the loss as a function of the amount of training data for two MNIST classification tasks~\cite{Howard:2024bbi}. 
The plot communicates three principal conclusions: the theoretical predictions agree closely with the empirical measurements, the losses exhibit semi power-law behavior, and the two classification tasks follow quantitatively different trends. 
Features such as the logarithmic axes, distinct colors for each task, error bars, and consistent representations of theory and empirical results make these conclusions visible without requiring the reader to reconstruct them from the surrounding text.

\subsection{Examples of Common Situations}
Here, we will cover some common scenarios one might encounter and possible ways to address them.
First, you will likely be in a situation where there are far too many things to plot. 
As a rule of thumb, plots will start to look too busy if there are more than 3-4 distinct lines. 
When this happens, ask which pairwise relationships are the most important to highlight to make the scientific point.
Then, try to split these comparisons into multiple plots or choose a representative comparison and say the rest in words.
See Fig.~\ref{fig:reduceLines} for an example of this. 

\begin{figure} [ht]
    \centering
    \includegraphics[width=0.9\linewidth]{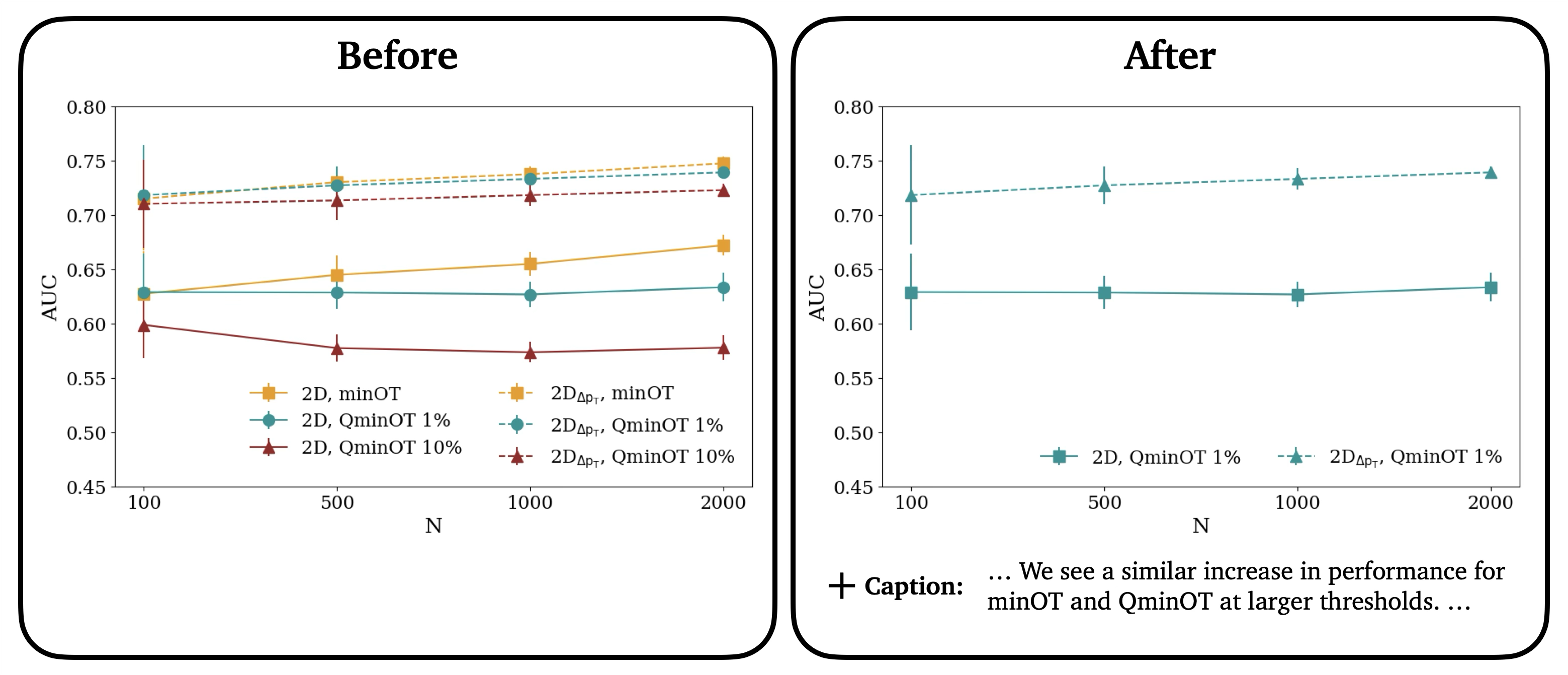}
    \caption{\textbf{Figure showing a potential fix for a plot with too many lines}. The primary goal of the plot is to show that ``2D'' methods perform worse than ``2D$_{\Delta {\rm p}_{\rm T}}$'' methods across all choices of $N$. However, the plot is trying to do too many pairwise comparisons at once. Since this performance gain is present for all methods (minOT, QminOT 1\%, and QminOT 10\%), we can simply choose a representative method (in this case QminOT 1\%) and say that we see a similar trend in the other methods in the caption. 
    If comparing the performance of the three methods is also important, one can show this in a separate plot.
    The data and methods for this example were taken from Ref.~\protect\cite{Craig:2024rlv}.
    }
    \label{fig:reduceLines}
\end{figure}

Another common situation is when you want to show the agreement between two quantities (for example theoretical prediction and empirical result) but the difference between them is much smaller than the overall shape.
In the case where both the shape and agreement are important, one can add a residual plot.
This residual plot shows the difference or ratio of the two quantities below the main plot.
See Fig.~\ref{fig:residualPlot} for an example.

\begin{figure}
    \centering
    \begin{subfigure}[b]{0.49\textwidth}
        \includegraphics[width=\textwidth]{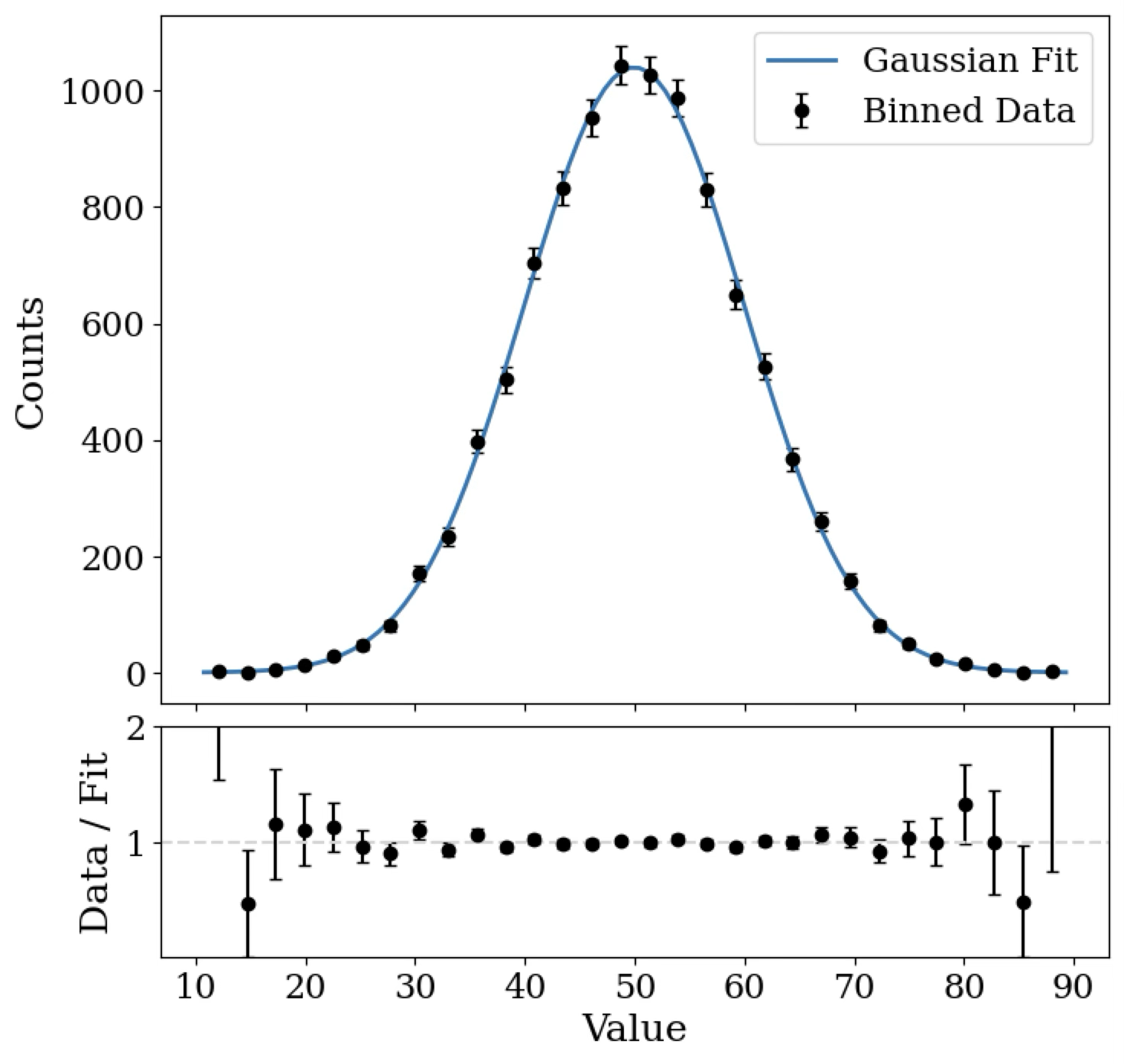}
        \caption{}
        \label{fig:ratioResidual}
    \end{subfigure}
    \begin{subfigure}[b]{0.49\textwidth}
        \includegraphics[width=\textwidth]{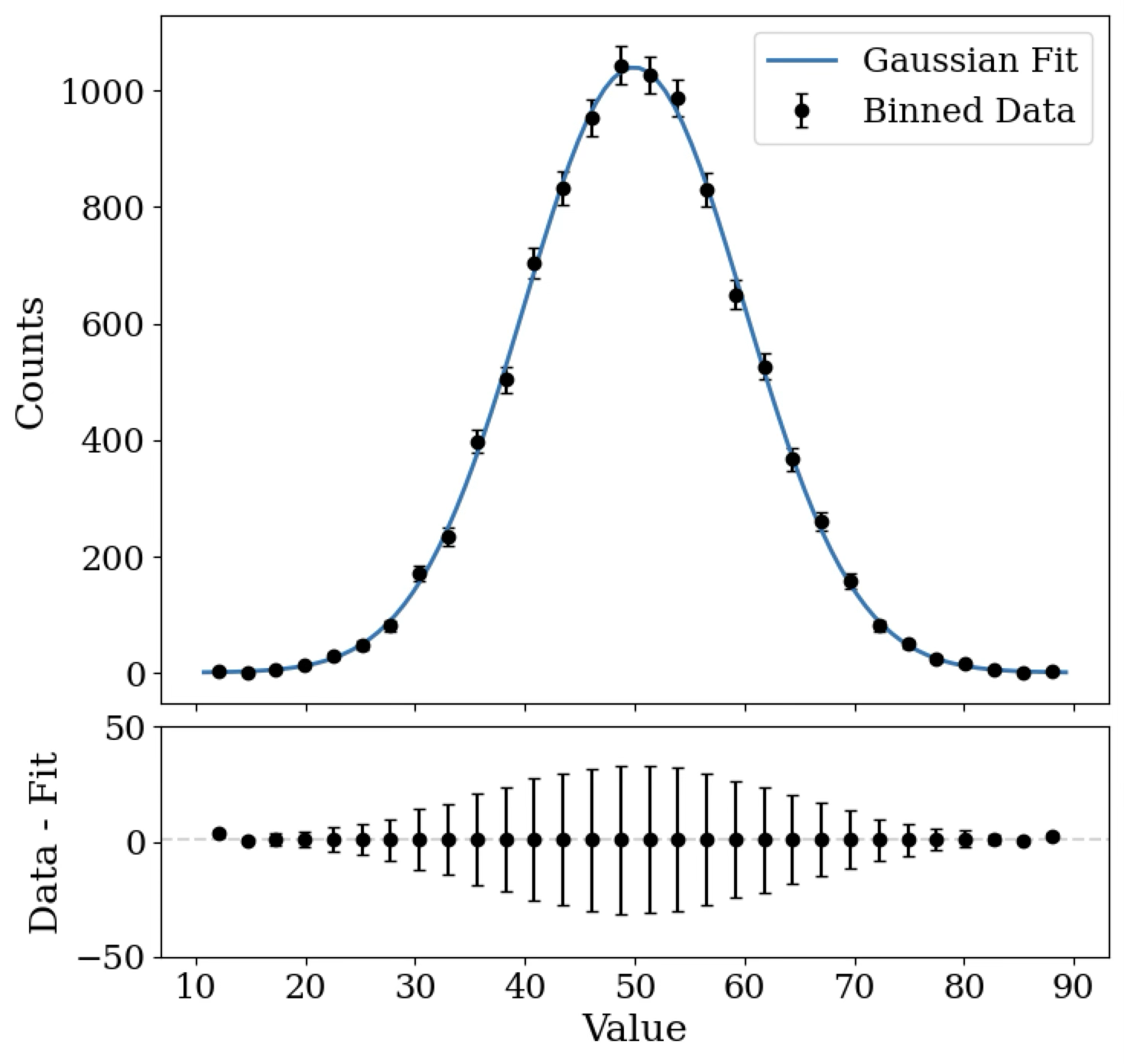}
        \caption{}
        \label{fig:diffResidual}
    \end{subfigure}
    \caption{\textbf{Figure showing two examples of residual plots}. When you want to show agreement between two quantities (in this case binned data and the best fit function) but also want to show the overall shape, it can be helpful to use a residual plot. In the main plot, we show the fitted function and binned data to give an idea of the agreement and shape of the distribution. However, more information about the fit can be gained by looking at the difference or ratio of the two quantities. This can be shown in a residual plot below the main plot. In this case, it helps accentuate differences happening at the tails of the distribution which would be difficult to notice otherwise.}
    \label{fig:residualPlot}
\end{figure}

\subsection{Plot Accessibility}

An important, but often neglected, aspect of plot design is accessibility. 
Labels, legends, and axis ticks should be large and clear enough to read at the figure’s final printed size.
Figures should include descriptive captions that explain not only what is displayed, but also the principal trends and conclusions the reader should draw from the plot. 
It is also important to note that the typical PDF files generated (e.g. by LaTeX) are not automatically compatible with accessibility tools such as screen readers. 
See Ref.~\cite{AccessiblePDFWorkflow} for an AI workflow to convert a PDF file into an accessible version. 

Color choices require particular care~\cite{Crameri2020} since plots should remain interpretable for readers with different forms of color vision deficiency (CVD).
Ideally, information should not be encoded through color alone; instead, distinct markers or line styles should be used to provide redundant visual cues.
Tools such as {\sc DaltonLens}~\cite{DaltonLens} and {\sc Coblis}~\cite{Coblis} can simulate different forms of CVD and help identify potential accessibility problems before publication.
Color maps should also be perceptually uniform so that features are not artificially accentuated.
Finally, because papers are still frequently printed in black and white, every figure should be checked in grayscale to ensure that its conclusions remain clear.

\newpage
\section{Coding Exercises}

In this section we describe the coding exercises included in the tutorial. 
The corresponding code notebooks are available at \faGithub~\href{https://github.com/jnhoward/Banff-Tutorial/}{jnhoward/Banff-Tutorial/}.
Brief solutions to the questions in the exercises are given in~\ref{app:ExerciseAnswers}.

\subsection{Exercise: Build Your Own ROC Curve}
Write a function that calculates a ROC curve and verify it against the built-in functions in the {\sc scikit-learn} library~\cite{scikit-learn}. Once you have finished, test your intuition by answering the following questions: (1) What score distributions would lead to an AUC of 0.5 (random guessing)? (2) What score distributions would lead to an AUC of 1 (perfect classifier)? (3) What is the meaning of an AUC that is less than 0.5?

\subsection{Exercise: Estimate the Uncertainty}\label{sec:Ex_estimateUncertainty}

Write a function that implements jackknife resampling.
For a set of mystery data, $X = \{x_1, \ldots, x_n \mid x_i \sim p(x) \}$, estimate the mean of $p(x)$ and the standard error on the mean.
Estimate these quantities three ways: (1) using separate samples, (2) using bootstrap resampling, (3) using your jackknife resampling function. 
How close are the methods to each other? 
To the true value?

\section*{Acknowledgments}
A huge thank you to Edward Hirst and Elli Heyes for organizing this session and encouraging us to write up our tutorials. 
And thanks to my fellow session leads Tomás S. R. Silva and Yidi Qi for helping make these tutorials a success.
JNH is supported by the National Science Foundation under Grant No. NSF PHY-2309135 and by a UCSB Chancellor's Postdoctoral Fellowship.

\section*{ORCID}
\noindent Jessica N. Howard - \url{https://orcid.org/0000-0001-8295-7727}

\appendix

\section{Exercise Solutions} \label{app:ExerciseAnswers}

\subsection{Build Your Own ROC Curve}

\begin{enumerate}
    \item \textbf{What score distributions would lead to an AUC of 0.5 (random guessing)?} If the score distributions of both classes are perfectly overlapped every threshold will have the same TPR and FPR. Thus, the ROC curve will be a straight line from $({\rm TPR}, {\rm TNR}) = (0,1)$ to $({\rm TPR}, {\rm TNR}) = (1,0)$ with an integrated area (AUC) of 0.5.
    \item \textbf{What score distributions would lead to an AUC of 1 (perfect classifier)?} If the score distributions of class \signal~and class \background~are separated by a gap and the \signal~scores are all greater than the \background~scores, 
    then any threshold in (or above) the gap will have ${\rm TNR} = 1$.
    \item \textbf{What is the meaning of an AUC that is less than 0.5?} This simply means that the order of the distributions is inverted (\background~ scores are greater than \signal~scores). 
    However, one can still effectively separate \signal~vs \background~with this classifier. 
    In the extreme case, we could have a gap between \signal~and \background, in which case the AUC is 0 but we have a perfect (inverse) classifier.
\end{enumerate}

\subsection{Estimate the Uncertainty}
The distribution $p(x)$ is a normal distribution with mean $\mu=1.4$ and variance $\sigma^2 = 4$.
The mystery data was generated by drawing $n=500$ samples.
Since $p(x)$ is Gaussian distributed we know theoretically that the true standard error on the mean should be $\frac{\sigma}{\sqrt{n}} = 0.0894$.
Thus, the correct answer is $\mu_0, \sigma_0 = 1.4, 0.0894$.

However, one will find that the estimates all predict an inaccurate $\hat{\mu}$ due to the small number of samples available. 
Using $N=5$ independent test sets has the largest uncertainty on its prediction. 
It predicts $\hat{\mu} = 1.2701 \pm 0.1208$, making the true value just slightly one standard deviation above this prediction.
Using $N=1000$ bootstrapped samples, the Bootstrap resampling strategy predicts $\hat{\mu} = 1.2712 \pm 0.0851$. 
It is more sure of its prediction which falls similarly far below the truth.
The Jackknife resampling strategy is the most overconfident, predicting $\hat{\mu} = 1.2701 \pm 0.0076$.

This illustrates the risks when dealing with limited data and indicates that $n=500$ is too little to accurately approximate $p(x)$.
Resampling this empirical distribution exacerbates this effect.
It also affirms that the most conservative estimates will come from using separate test sets.
Thus, this is the primary strategy used in ML results, but it relies on the fact that there is typically plenty of independent data.

\bibliographystyle{ws-ijdsms}
\bibliography{references}

% <<< END PASTE: TUTORIAL 3 DOCUMENT BODY <<<

\end{tutorial}

\end{document}